%% file: main.tex
\PassOptionsToPackage{final}{changes}
\documentclass[12pt]{elsarticle}

\usepackage[utf8]{inputenc}
\usepackage{microtype}
\usepackage{subfigure}
\usepackage{booktabs} % for professional tables
\usepackage{amsfonts,amsmath,amsthm,amssymb}
\usepackage{bbm}
\usepackage{dsfont}
\usepackage{tikz}
\PassOptionsToPackage{hyphens}{url}\usepackage{hyperref}
\usepackage{pgfplots}
\usepgfplotslibrary{fillbetween}
\pgfplotsset{width=0.8\textwidth,compat=newest}
\usepackage{enumitem}
\usepackage{algorithm}
\usepackage{algorithmic}
\usepackage{standalone}
\usepackage{url}
\usepackage[acronym]{glossaries}
\usepackage{changes}

\usetikzlibrary{dsp,chains,calc}

\usepackage[capitalize]{cleveref}

\def\Alg{\mathcal A}

\def\FFF{\mathcal F}
\def\XXX{\mathcal X}
\def\YYY{\mathcal Y}
\def\SSS{\mathcal S}
\def\TTT{\mathcal T}
\def\NNN{\mathcal N}

\def\DecisionSet{\widehat{\mathcal{T}}}

\newcommand{\vt}[1]{\mathbf{#1}}
\newcommand{\ol}[1]{\overline{#1}}
\newcommand\norm[1]{\left\lVert#1\right\rVert}
\def\RR{\mathbb{R}}
\def\EE{\mathbb{E}}
\def\Noise{W}
\def\RDI{J}
\newcommand{\Var}{\mathrm{Var}}

\DeclareMathOperator*{\esssup}{ess\,sup}
\DeclareMathOperator*{\essinf}{ess\,inf}
\DeclareMathOperator*{\argmax}{arg\,max}
\DeclareMathOperator*{\argmin}{arg\,min}

\def\Remp{\mathcal R_{\mathrm{emp}}}
\def\Rexp{\mathcal R_{\mathrm{exp}}}
\def\ellmax{\ell_{\mathrm{max}}}
\def\Rmax{R_{\mathrm{max}}}
\def\RSG{R_\mathrm{SG}}
\def\EG{\mathcal E_{\mathrm{G}}}
\def\GG{\mathcal G_{\mathrm{G}}}
\def\PS{\mathcal P_{\mathrm{Suc}}}
\def\Prob{\mathrm P}
\def\Tr{\mathrm{Tr}}
\def\TrainSet{\vt{z}}

\def\SI{S}
\def\TI{T}
\newcommand{\smallparameter}{\epsilon}

\newtheorem{theorem}{Theorem}
\newtheorem{lemma}{Lemma}
\newtheorem{proposition}{Proposition}
\newtheorem{definition}{Definition}

\def\BibTeX{{\rm B\kern-.05em{\sc i\kern-.025em b}\kern-.08em
    T\kern-.1667em\lower.7ex\hbox{E}\kern-.125emX}}

\crefname{equation}{Equation}{Equations}
\Crefname{equation}{Equation}{Equations}
\crefformat{equation}{#2(#1)#3}
\crefrangeformat{equation}{#3(#1)#4--#5(#2)#6}
\crefmultiformat{equation}{#2(#1)#3}{ and~#2(#1)#3}{, #2(#1)#3}{ and~#2(#1)#3}
\Crefformat{equation}{#2(#1)#3}
\Crefrangeformat{equation}{#3(#1)#4--#5(#2)#6}
\Crefmultiformat{equation}{#2(#1)#3}{ and~#2(#1)#3}{, #2(#1)#3}{ and~#2(#1)#3}

\crefname{appendix}{}{}
\Crefname{appendix}{}{}

\glsdisablehyper

\journal{Neurocomputing}

\begin{document}

\include{includes/definitions}

\begin{frontmatter}

  \title{Bounding Information Leakage in Machine Learning}

  \author[Inria,Ecole]{Ganesh Del Grosso}

  \author[TUWien]{Georg Pichler}

  \author[Inria,Ecole]{Catuscia Palamidessi}

  \author[ILLS]{Pablo Piantanida}

  \affiliation[Inria]{organization={Inria~Saclay, team~COMETE},
    addressline={1~Rue~Honore~d'Estienne~d'Orves}, 
    city={{} Palaiseau},
    postcode={91120}, 
    state={Ile-de-France},
    country={France}}
  
  \affiliation[Ecole]{organization={Ecole~Polytechnique, LIX},
    addressline={1~Rue~Honore~d'Estienne~d'Orves}, 
    city={{} Palaiseau},
    postcode={91120}, 
    state={Ile-de-France},
    country={France}
  }

  \affiliation[TUWien]{organization={TU~Wien, Institute~of~Telecommunications},
    addressline={{} Gusshausstrasse~25/E389}, 
    % city={Vienna},
    postcode={1040}, 
    state={Vienna},
    country={Austria}}

  \affiliation[ILLS]{organization={International Laboratory on Learning Systems (ILLS)\\
      Universite McGill - ETS - MILA - CNRS - Universite Paris-Saclay - CentraleSupelec},
    addressline={1100 Rue Notre Dame O}, 
    city={Montreal},
    postcode={H3C 1K3},
    state={Quebec},
    country={Canada}}
  
  % \affiliation[ILLS]{organization={International Laboratory on Learning Systems (ILLS) Université McGill - ETS - MILA - CNRS - Université Paris-Saclay - CentraleSupélec},
  % addressline={1100 Rue Notre Dame O},
  % city={Montréal},
  %   postcode={H3C 1K3},
  %   state={Quebec},
  %   country={Canada}}

  \begin{abstract}
    \input{abstract}
  \end{abstract}

  % \begin{highlights}
  % \item Research highlight 1
  % \item Research highlight 2
  % \end{highlights}

  \begin{keyword}
    Membership Inference, Privacy, Attacks in Machine Learning.
  \end{keyword}

\end{frontmatter}

\section{Introduction}
\input{introduction}

\section{Preliminaries}
\label{sec:Prelim}
\input{preliminaries}

\section{Main Results}
\input{mainresults}

\section{Examples and Numerical Experiments }
\label{sec:Exp}
\input{experiments}

\section{Summary and Concluding Remarks}
\label{sec:Conclusion}
\input{conclusion}

\newpage
\appendix

% \begin{center}
%   {\LARGE\textbf{\textsc{Appendix}}}  
% \end{center}
\input{appendix}

\section*{Acknowledgment}

This research was supported by DATAIA ``Programme d'Investissement d'Avenir'' (ANR-17-CONV-0003) and by the ERC project Hypatia under the European Union's Horizon 2020 research and innovation program, grant agreement No.\ 835294.

\newpage
\bibliographystyle{elsarticle-num}
\bibliography{Bib}

\end{document}

%% file: includes/definitions.tex
\pgfplotstableset{col sep=comma}

\newlength{\standaloneWidth}
\newlength{\standaloneHeight}
\setlength{\standaloneWidth}{.85\textwidth}
\setlength{\standaloneHeight}{4.8cm}

\pgfplotsset{
  basicAxis/.style={
    major grid style={line width=.2pt,draw=gray!50},
    grid=both,
    width = \linewidth,
    height = .6\linewidth,
    x label style ={yshift=-\baselineskip, anchor=south},
    y label style = {yshift=0.0cm, anchor=south},
    ytick={0.6,0.7,0.8},
    yticklabels={.6,.7,.8},
    ymin = 0.5, ymax = .9,
    xmin=10, xmax=20000,
    legend style={nodes={scale=0.7, transform shape}, at={(0.03,0.95)},anchor=north west},
  },
  leftAxis/.style={
    basicAxis,
    axis y line*=left,
    xlabel={$n$},
    ylabel style = {align=center},
    ylabel={Success Rate},
  },
  rightAxis/.style={
    basicAxis,
    axis y line*=right,
    ylabel style = {align=center},
    ylabel={Accuracy},
    y label style = {yshift=-0.5cm, anchor=south},
    xtick align=inside,
  },
  leftAxis3/.style={
    major grid style={line width=.2pt,draw=gray!50},
    grid=both,
    ymin = 0.45, ymax = 1,
    ytick={.5,.7,.9},
    yticklabels={.5,.7,.9},
    width = \standaloneWidth,
    height = \standaloneHeight,
    axis y line*=left,
    xlabel={$n$},
    ylabel style = {align=center},
    ylabel={Success Rate},
    xtick align=inside,
  },
  rightAxis3/.style={
    major grid style={line width=.2pt,draw=gray!50},
    grid=both,
    ymin = 0.45, ymax = 1,
    ytick={.5,.7,.9},
    yticklabels={.5,.7,.9},
    width = \standaloneWidth,
    height = \standaloneHeight,
    axis y line*=right,
    ylabel style = {align=center},
    ylabel={Accuracy},
    y label style = {yshift=-0.5cm, anchor=south},
  },
  AttAxis/.style={
    major grid style={line width=.2pt,draw=gray!50},
    grid=both,
    width = \standaloneWidth,
    height = \standaloneHeight,
    xlabel={$n$},
    xtick align=inside,
  },
  rightAxis1/.style={
    basicAxis,
    axis y line*=right,
    ylabel style = {align=center},
    y label style = {yshift=-0.5cm, anchor=south},
    ytick={.625,.75,.875},
    yticklabels={.625,.75,.875},
    ymin = 0.5, ymax = 1,
    xmin = 20, xmax=141,
  },
  leftAxis1/.style={
    basicAxis,
    axis y line*=left,
    xlabel={$n$},
    ylabel style = {align=center},
    ylabel={Success Rate},
    ytick={0.25,.5,.75},
    yticklabels={0.25,.5,.75},
    ymin = 0, ymax = 1,
    xmin = 20, xmax=141,
  },
  axis1/.style={
    basicAxis,
    ytick={.5,.6,.7},
    yticklabels={.5,.6,.7},
    xlabel= {$n$},
    ylabel= {Success Rate.},
    xmin = 20, xmax=141,
    ymin = 0.45, ymax = 0.8,
  },
  table/basicTable/.style={
    col sep=comma
  },
  axis2/.style={
    basicAxis,
    ytick={.5,.7,.9},
    yticklabels={.5,.7,.9},
    xlabel= {$n$},
    ylabel= {Success Rate.},
    xmin = 20, xmax=141,
    ymin = 0.45, ymax = 1,
  },
  table/basicTable/.style={
    col sep=comma
  },
  table/table1/.style={
    basicTable,
    x = a,
    y = b,
  },
  plotDef/.style={
    mark=none,
  },
  scatterDef/.style={ %scatter,
    % only marks
  },
  attInfDef/.style={
    xmin=90, xmax=10000,
    xtick align=inside,
    xtick={100, 200, 300, 400, 500, 600, 700, 800, 900, 1000, 2000,
      3000, 4000, 5000, 6000, 7000, 8000, 9000, 10000},
    xticklabels={$10^2$, , , , , , , , , $10^3$, , , , , , , $8\cdot 10^3$},
  },
}

\newacronym{ml}{ML}{Machine Learning}
\newacronym{mia}{MIA}{Membership Inference Attack}
\newacronym{dp}{DP}{Differential Privacy}
\newacronym{pdf}{pdf}{probability density function}
\newacronym{pmf}{pmf}{probability mass function}
\newacronym{dnn}{DNN}{Deep Neural Network}

%% file: abstract.tex
Recently, it has been shown that Machine Learning models can leak sensitive information about their training data. This information leakage is exposed through membership and attribute inference attacks. Although many attack strategies have been proposed, little effort has been made to formalize these problems. We present a novel formalism, generalizing membership and attribute inference attack setups previously studied in the literature and connecting them to memorization and generalization. First, we derive a universal bound on the success rate of inference attacks and connect it to the generalization gap of the target model. Second, we study  the question of how much sensitive information is stored by the algorithm about its training set and we derive bounds on the mutual information between the sensitive attributes and model parameters. Experimentally, we illustrate the potential of our approach by applying it to both synthetic data and classification tasks on natural images. Finally, we apply our formalism to different attribute inference strategies, with which an adversary is able to recover the identity of writers in the PenDigits dataset. 

%% file: introduction.tex
\Gls{ml} models have been known to leak information about their training records. This raises severe privacy concerns in cases where the training data contains sensitive information, for instance, when using real patients' data in medical applications, e.g., \cite{LUNDERVOLD2019102,Katzman_2018}.
In order for an \gls{ml} algorithm to be private, the General Data Protection Regulation (GDPR) and similar laws require for it to be impossible to \emph{single out} any individual from the training set. Recently, \cite{DBLP:journals/corr/abs-2005-08679} pointed out the need for a clear definition of what this means. Nonetheless, it is widely considered in the literature that \glspl{mia} measure the privacy of \gls{ml} algorithms, i.e., if a \gls{mia} is effective against a model, it is possible to single out an individual from its training set.

Membership Inference has been studied in the literature, but the efforts have been concentrated on model or data dependent strategies to perform the attacks, rather than developing a general framework to understand these problems. This drives us to propose a novel formalism, providing a general framework to study inference attacks and their connection to generalization and memorization. Compared to previous works (e.g. \cite{DBLP:journals/corr/abs-1709-01604, pmlr-v97-sablayrolles19a}), we consider a more general framework, where we study the performance of the Bayesian attacker without making any assumptions on the
distribution of model parameters given the training set.

Furthermore, the present paper is distinguished from previous work by studying the connection between the information stored by the model and the leakage of sensitive information. Intuitively, the amount of leaked information should be proportional to the amount of training data stored by the model. In our study, we find that the mutual information between training set and model parameters upper bounds the gain of the Bayesian attacker over an attacker that only uses the prior distributions of the sensitive attributes.

Lastly, we consider the risk of sensitive information leakage from \gls{ml} models in the form of attribute inference attacks. In these attacks, having partial knowledge of a sample in the training set, an adversary tries to extract sensitive information about the sample from the target model. In the case of medical data, the sensitive information could be the genetic profile of a patient \cite{184489}. We study several attribute inference strategies against \gls{ml} models. Our framework allows us to formalize these problems, draw universal bounds on the performance of \glspl{mia} and find connections between generalization and privacy.

\subsection{Summary of Contributions}

Our work investigates  fundamental bounds on information leakage and advances the state-of-the-art in multiple ways.
\begin{enumerate}[wide, labelindent=0pt, label=\textbf{\arabic*.}]

\item \emph{A simple framework for modeling membership and/or attribute inference attacks.} We introduce a probabilistic framework for the analysis of membership and/or attribute inference attacks on machine learning systems. The necessary \cref{def:success_prob,def:att_inference,def:mem_inference} are simple and concise, yet flexible enough to be applied to different problem setups. 
% Based upon this framework, we draw formal conclusions, which we validate experimentally.

\item \emph{Universal bounds on the success rate of inference attacks}.
By considering the success rate obtained by the Bayes decision rule, we are able to draw strong conclusions about the privacy of a \gls{ml} model. The attacker we consider is given with perfect knowledge of the underlying probability distribution. As such, it provides an upper bound for the probability of success of any attack strategy (\cref{thm:neyman_pearson_optimality}). As a matter of fact, this bound represents a privacy guarantee for any \gls{ml} model and may be useful to guide the design of privacy defense mechanisms.

\item \emph{Relation between generalization gap and membership inference.} A model that does not generalize well is susceptible to \glspl{mia}. \Cref{thm:bounded_loss}, which generalizes \cite[Theorem~1]{DBLP:journals/corr/abs-1709-01604}, provides a lower bound for the case of bounded loss functions, while \cref{thm:MSE_gaussian,thm:MSE_Exp} cover the case of sub-Gaussian and tail-bounded loss functions, respectively. These results provide formal evidence that bad generalization leads to privacy leakage. However, the converse does not hold in general, i.e., \emph{good generalization does not automatically prevent privacy leakage}. Intuitively, one would think that a model that generalizes well would be agnostic to any particular sample being in its training set. Nevertheless, we show, by providing a suitable counterexample, that this intuition is wrong.

\item \emph{Missing information in inference attacks and its connection to generalization.} Using mutual information, we study the amount of information stored by a trained model about its training set, and the role this information plays when the model is susceptible to privacy attacks (\cref{the4}). We find that the mutual information between the sensitive attribute and the model parameters upper bounds the gain of the Bayesian attacker over an attacker that uses the prior distribution of the sensitive attribute. This mutual information is in turn upper bounded by the mutual information between the training set and the trained model parameters.

\item \emph{Numerical experiments.} As proof of concept we consider linear regression with Gaussian data (\cref{sec:gaussians}). The simplicity of this setup allows us to estimate the success rate of the Bayesian white-box attacker and, since the loss is exponentially tail-bounded, we can also apply \cref{thm:MSE_Exp} to monitor the interplay between success rate and generalization. Then we apply our theoretical results in a more practical setting; namely, \glspl{dnn} for classification (\cref{sec:DNNs}). Considering bounded loss functions, we apply \cref{thm:bounded_loss} to lower bound the success probability of the Bayesian attacker. We perform \glspl{mia} using state of the art strategies to assess the quality of this bound. Lastly, to illustrate that a model susceptible to membership inference might be susceptible to other, more severe, privacy violations, we consider a model for hand-written digit classification and use this example to apply several attribute inference strategies, comparing their effectiveness (\cref{sec:pendig}).
\end{enumerate}

\subsection{Related Work}

\textbf{Connection between Privacy Leakage and Generalization:} \cite{DBLP:journals/corr/abs-1709-01604} study the interplay between generalization, \gls{dp}, attribute and membership inference attacks. Our work investigates related questions,  but offers a different and complementary perspective. While their analysis considers only bounded loss functions, we extend the results to the more general case of tail-bounded loss functions. They consider a membership inference strategy that uses the loss of the target model, yielding an equivalence between generalization gap and success rate of this attacker. In contrast, we consider a Bayesian attacker with white-box access, yielding an upper bound on the probability of success of all possible adversaries and also on the generalization gap.

Consequently, a large generalization gap implies a high success probability for the attacker. The converse statement, i.e., ``\emph{generalization implies privacy}" has been proven false in previous works, such as \cite{236216,DBLP:journals/corr/abs-1802-04889,DBLP:journals/corr/abs-1709-01604}. Our work also provides a counter proof, giving an example where the generalization gap tends to $0$ while the attacker achieves perfect accuracy.

In this line of work, \cite{pmlr-v97-sablayrolles19a} derived an attack strategy for membership inference that is optimal to their setup. However, their results rely on randomness during training and assume a specific form in the distribution of network parameters given the training set. In this sense, our Bayesian attacker can be specialized to their framework and models. 

\added{The authors of concurrent work~\cite{TanTradeoff} studied the trade-off between the size of the target model (number of model parameters) and the success rate of an optimal attacker within their framework. That setup differs from ours mainly in terms of the capabilities of the attacker; while our attacker has access to the model parameters and full information on the target sample, their attacker only has access to the target sample data and corresponding model output. The work~\cite{TanTradeoff} presents a formal relation between the over-parametrization of the model and the success rate the Bayesian attacker against a linear regression model trained on Gaussian data. Differences in the definition of the sample-space, target model and attacker capabilities lead to orthogonal results, but similar conclusions.
}

\textbf{Membership Inference:} \cite{DBLP:journals/corr/ShokriSS16} utilize \glspl{mia} to measure privacy leakage in deep neural networks. Their attacks consist in training a classifier that distinguishes members from non-members. While their first work covers the case of black-box attacks, subsequent work by \cite{Nasr_2019} considers white-box attacks, where the adversary has access to the model parameters. Later, \cite{8634878} studied the influence of model choice on the privacy leakage of \gls{ml} models via membership inference. 

Recent works \cite{Jayaraman2021RevisitingMI,Rezaei,Song2021,Carlini2022} revise new and old membership inference strategies under the light of new evaluation metrics. In particular, the work in \cite{Carlini2022} takes inspiration from \cite{pmlr-v97-sablayrolles19a}, developing an attack strategy based on estimating the distribution of the loss. \added{Further work \cite{StolenLeino} proposes to use learned differences in distribution between outputs of intermediate layers to predict membership.} In \cite{MIAAE}, a new \gls{mia} strategy is proposed, which is based on the magnitude of the perturbation necessary to successfully make the target model change its prediction. It is compared to state-of-the-art methods \cite{Rezaei,Nasr_2019}.

\added{The use of shadow models is prevalent in the \gls{mia} literature. These models mimic the behavior of the target model, while allowing an attacker access to the training set and model parameters. Many of the aforementioned \glspl{mia} require the training of an attacker model (e.g.\ \cite{Rezaei}), while others require the training of shadow models \cite{DBLP:journals/corr/ShokriSS16,Nasr_2019,StolenLeino,SalemMLeaks} in addition to training an attacker. The attacks in \cite{Song2021} require only black box access to the model and no additional information, while the attacks in \cite{MIAAE} require white box access.}

\added{Recent work \cite{ChenPoison}, applies the Modified Entropy strategy proposed by \cite{Song2021} to launch \glspl{mia} against poisoned target models. This setup differs from previous works in the sense that the attacker plays an active role in the training by poisoning part of the training data. \cite{ChenPoison} shows that the effectiveness of \glspl{mia} is highly increased against poisoned models.
}

Typically, when studying the privacy leakage of \gls{ml} models, classifiers are considered as the target to privacy attacks. In contrast, \cite{LOGANHayes} were the first to consider \glspl{mia} against generative models. A comprehensive study of \glspl{mia} against GANs and other generative models is provided by \cite{GANleaksChen}.

\textbf{Attribute Inference/Model Inversion:} A more severe violation of privacy is represented by attribute inference attacks. Mainly two forms of these attacks have been considered in the literature. The first consists in inferring a sensitive attribute from a partially known record plus knowledge of a model that was trained using this record, e.g. \cite{184489,8476925,toomuchSong,feasibilityZhao,MelisExploiting,QueryEfficient}. The second consists in generating a representative sample of one of the members of the training set, or one of the classes in a classification problem, by exploiting knowledge of the target model, e.g. \cite{10.1145/2810103.2813677,DBLP:journals/corr/abs-1910-04257,Baumhauer2020MachineUL,10.1145/3319535.3354261,hitaj2017deep,UpdatesLeak}. Our framework is applicable to both forms, but in this work we focus on the former, i.e., inferring sensitive information from a partially known record. \cite{7536387} propose a framework that generalizes to both types of attribute inference attacks and connects them to several cryptographic notions. The notion of attribute inference is also formalized by \cite{DBLP:journals/corr/abs-1709-01604}. While their work defines the advantage of an adversary as the difference between the information leaked by the model and the information present in the  underlying probability distribution of the data, our formalism only allows the adversary to gain advantage from the target model. Furthermore, we consider and compare different attack strategies, while their work only focuses on the attack introduced by~\cite{184489}, and an attacker with oracle access to a membership inference algorithm. 

\textbf{Model Extraction:} A third class of privacy violation consists in stealing the functionality of a model, when the model and its parameters are considered sensitive information, e.g., \cite{10.5555/3241094.3241142}, but this setup is out of the scope of our work.

\textbf{Unintended Memorization:} Leakage of sensitive information might be caused by unintended memorization by the model. \cite{236216} studies unintended memorization by generative sequence models. They prove that unintended memorization is persistent and hard to avoid; moreover, they find that a model can present exposure even before overfitting. This is an instance in which a model can leak sensitive information even while generalizing well.

\textbf{Differential Privacy in Machine Learning:} \Gls{dp}, \cite{10.1007/11787006_1, 10.1561/0400000042} is a widely used definition of privacy, which guarantees the safety of individuals in a database while releasing general information about the group. There have been several works in \gls{ml} that use \gls{dp} as a measure for privacy or use \gls{dp} mechanisms for defense against inference attacks. \cite{10.1145/2976749.2978318} proposes a Differentially Private Stochastic Gradient Descent method for training neural networks. Their analysis allows them to estimate the privacy budget when successively applying noise to the model parameters during training. Later, \cite{Zhao_2020} presented a comprehensive analysis of \gls{dp} in \gls{ml} by considering the different stages in which noise can be added to make an \gls{ml} model differentially private. \cite{236254} evaluates the effectiveness and cost of \gls{dp} methods for \gls{ml} in the light of inference attacks. \cite{DBLP:journals/corr/abs-1901-09697} propose \emph{Bayesian \gls{dp}}, which takes into account the data distribution to provide more practical privacy guarantees, achieving the same accuracy as \gls{dp} while providing better privacy guarantees on several models and datasets. \added{Recent work~\cite{auditLu} proposes an algorithm to ``audit'' the privacy of \gls{ml} models, accurately computing the privacy budget necessary to prevent attacks with minimal impact on the utility of the target model.} We do not consider the connection between \gls{dp} and \glspl{mia}, as this is thoroughly analyzed in \cite{DBLP:journals/corr/abs-1709-01604}.

\textbf{Federated Learning:} Inference attacks that target federated systems have been investigated by \cite{hitaj2017deep,bagdasaryan2019backdoor}. Privacy preserving methods specific for federated learning have been proposed by \cite{so2020scalable,10.1145/3133956.3133982,7447103,kone2016federated}. \cite{surveyFL} provides a comprehensive study of \glspl{mia} against Federated Learning models. In these setups the attacker can influence other entities during training. In our framework the attacker directly obtains the trained model; thus, our framework does not cover such cases.

\textbf{Adversarial Examples and Privacy:} There have been several works that combine the topics of privacy and adversarial examples. \cite{Song_2019} studies the impact that securing  a Machine Learning model against Adversarial Attacks has on the privacy of the model. \cite{DBLP:journals/corr/abs-1909-10594} makes use of Adversarial Examples as part of a defense mechanism against \glspl{mia}. \cite{phan2020scalable} were the first to simultaneously address the issues of robustness and privacy, providing a complete analysis of both aspects of \glspl{dnn}.
%%% Local Variables:
%%% mode: latex
%%% TeX-master: "main"
%%% End:

%% file: preliminaries.tex
A random variable is indicated by upper case (e.g., $X$). Lower case letter indicate realizations, while calligraphic case denotes the alphabet (e.g., $X \sim \XXX$ and $x \in \XXX$). 
\added{A \gls{pdf} is denoted by $p$ (e.g., the \gls{pdf} of $X$ is denoted by $p_X$).}
Expectations $\EE[\cdot]$ are taken over all random variables inside the square brackets.

We assume a fully Bayesian framework, where $Z = (X,Y) \sim p_{XY} \equiv  p_Z$ denotes data $X$ and according labels $Y$, drawn from sets $\XXX$ and $\YYY$, respectively. The training set consists of $n$ i.i.d.\ copies $\TrainSet \triangleq \{z_1,\dots , z_n\}$ drawn according to  $\vt Z \sim p_Z^n$.

\subsection{Learning and Inference}

Let $\FFF \triangleq \left\{ f_\theta\, | \, \theta \in \Theta\right\}$ be a \emph{hypothesis  class} of (possibly randomized) decision functions  parameterized with $\theta$, i.e., for every $\theta \in \Theta$, $ f_\theta({}\cdot{};x) $ is a probability distribution on $\YYY$. We will abuse notation and let $f_\theta(y;x)$ be a \gls{pmf} or a \gls{pdf} in $y$ for every $x \in \XXX$, depending on the context.
The symbol $\hat Y_\theta(x)$ will be used to denote the random variable on $\YYY$ distributed according to $f_\theta( {}\cdot{};x)$.
In case the decision functions are deterministic, i.e., $f_\theta(y;x) \in \{0,1\}$ is a one-hot \gls{pmf} for every $\theta \in \Theta$, $x \in \XXX$, we write $\hat y_\theta(x) \in \YYY$ to denote this deterministic decision, i.e., $\hat y_\theta(x)=\argmax_{y\in\mathcal{Y}} f_\theta(y;x)$.

A \emph{learning algorithm} is a (possibly randomized) algorithm $\Alg$ that assigns to every training set $\TrainSet \in \left(\XXX \times \YYY\right)^n$ a probability distribution on the parameter space $\Theta$ (and, thus, also on the hypothesis space $\FFF$). We have $\Alg\colon \TrainSet \mapsto \Alg({}\cdot{};\TrainSet)$, where $\Alg({}\cdot{};\TrainSet)$ is a probability distribution on $\Theta$. The symbol $\widehat\theta(\TrainSet)$ is used to denote a random variable on $\Theta$, distributed according to $\Alg({}\cdot{};\TrainSet)$.
In case of a deterministic learning algorithm, we have a \gls{pmf} $\Alg(\theta ;\TrainSet) \in \{0,1\}$ for every training set $\TrainSet$ and can thus define the function $\widehat\theta(\TrainSet) = \argmax_{\theta \in \Theta} \Alg(\theta ;\TrainSet)$, yielding the (possibly random) decision function $f_{\widehat\theta(\TrainSet)}$.

To judge the quality of a decision function $f \in \FFF$ we require a loss function $\ell\colon \YYY \times \YYY \to \RR$. We naturally extend this definition to vectors by an average over component-wise application, i.e., $\ell(\vt y, \vt y') = \frac{1}{n} \sum_{i=1}^n \ell(y_i, y'_i)$.
\begin{definition}[Expected risks]
  \label{def:expected_risks}
  \added{We define $\varrho(\theta, (x,y)) \triangleq \EE[\ell(\hat Y_{\theta}(x), y)]$ as the expected loss between $f_\theta(x)$ and $y$. This notation is naturally extended to vectors as
    \begin{align}
      \varrho(\theta, \mathbf{z}) \triangleq \frac1n \sum_{i=1}^n \EE[\ell(\hat Y_{\theta}(x_i), y_i)] .
    \end{align}
    The \emph{expected risk} and \emph{empirical risk} of a learning algorithm $\Alg$ at training set $\mathbf Z$ are defined respectively as}\footnote{Note that the expectation is taken over all random quantities, i.e., $\mathbf Z \sim p_Z^n$, $\widehat\theta(\mathbf Z) \sim \Alg({}\cdot{};\mathbf Z)$, and $(X,Y)\sim p_Z$.} %\footnote{Note that the empirical risk is computed using the training data of the algorithm.}
  \added{
    \begin{equation}
      \Rexp(\Alg) \triangleq \EE\big[ \varrho\big (\widehat\theta(\mathbf Z), (X, Y)\big) \big] \;,\; \Remp(\Alg) \triangleq \varrho\big (\widehat\theta(\mathbf Z), \mathbf Z \big) \;,
    \end{equation}
    where the training set $\mathbf Z$ and $(X, Y)$ are independent.
    The difference between expected and empirical risk is the \emph{generalization gap} $\GG(\Alg)$, and its expectation $\EG(\Alg)$, which are respectively defined as}
    \begin{equation}
      \GG(\Alg) \triangleq \Rexp(\Alg) - \Remp(\Alg) \;,\; \EG(\Alg) \triangleq \EE[\GG(\Alg)] \,. \label{eq-generalization-error}
    \end{equation}
  \end{definition}

\subsection{Attack Model and Assumptions}

In order to make privacy guarantees for an algorithm $\Alg$, we need to specify an attacker model and the capabilities of an attacker. We will adopt a point of view of information-theoretic privacy and will not make assumptions about the computation power afforded to an attacker. We will also assume that the attacker has perfect knowledge of the underlying data distribution $p_Z$, as well as the algorithm $\Alg$.

In general, the goal of the attacker is to infer some property of $\TrainSet$ from $\widehat\theta(\TrainSet)$. However, in general the attacker may have access to certain side information. This may include the specific potential member of the training set that is queried (in case of a \gls{mia}) or any additional knowledge gained by the attacker. This side information is modeled by a random variable $\SI\in\SSS$, dependent on $\vt Z$, the value of which is known to the attacker. The attacker is interested in a target (or concept) property denoted by a random variable $\TI\in \TTT$, which is  also dependent on $(\vt Z, \SI)$. A (white box) \emph{attack strategy} is a (measurable) function $\varphi\colon \Theta \times \SSS \to \TTT$.

We shall assume that $\SI$ and $\TI$ are independent, but not necessarily conditionally independent given $\vt Z$. This natural assumption ensures that knowledge of the side-information $\SI$ does not change the prior $p_T = p_{T|S}$ of the attacker.

\begin{definition}
\label{def:success_prob}
  The Bayes \emph{success probability} of a (randomized) attack strategy $\varphi$ is
  \begin{align}
    \PS(\varphi) = \Prob\{\varphi(\widehat\theta(\vt Z), S) = T \}.
  \end{align}
  We may additionally define the \emph{success probability conditioned on side information $S = s$} as
  \begin{align}
    \PS(\varphi|s) = \Prob\{\varphi(\widehat\theta(\vt Z), s) = T | \SI = s\}.
  \end{align}
\end{definition}

\begin{definition}[Attribute inference attack]
\label{def:att_inference}
    We model the non-sensitive attribute by a random variable $V\in\mathcal{V}$. In this context, the input to the model is formed by the sensitive and non-sensitive attributes $X\equiv (V,T)$. Thus $\mathcal{X}\subseteq\mathcal{V}\times\mathcal{T}$. The side information given to the attacker can consist of $S=V$ or $S=(V,Y)$, depending on the attack strategy considered. In this work we only consider attribute inference attacks were $\TTT$ is finite.
\end{definition}

\begin{definition}[\gls{mia}]
\label{def:mem_inference}
  \label{def:memb_inf}
  In a \gls{mia}, let $T$ be a Bernoulli variable on $\mathcal{T} = \{0,1\}$ and $J$ is independently, uniformly distributed on $\{1,2,\dots,n\}$. Then set $S = TZ_J + (1-T)Z$, where $Z_J$ is a random element of the training set and $Z \sim p_Z$ is independently drawn. Thus, an attacker needs to determine if $T=1$, i.e., whether $S$ is part of the training set or not.

  \added{For later use we define the random variable $R \triangleq \varrho(\widehat\theta(\vt Z), S)$, i.e., the (random) loss function evaluated at $S$ (cf.\ \cref{def:expected_risks}).}
\end{definition}

\begin{figure}
    \centering
    \input{Images/mia.tikz}
    \caption{Scheme of a \gls{mia}. If $t=1$, the target sample is drawn from the training set $\mathbf z = (z_1, \dots, z_n)$ used by $\mathcal A$ to train the target model. If $t=0$, the target sample is independently drawn from the data distribution. The attacker $\varphi$ then uses the parameters $\theta$ at the output of $\mathcal A$ and the side information $s$ to provide an estimate $\hat t$ of $t$.}
    \label{fig:ArbitraryMIA}
\end{figure}
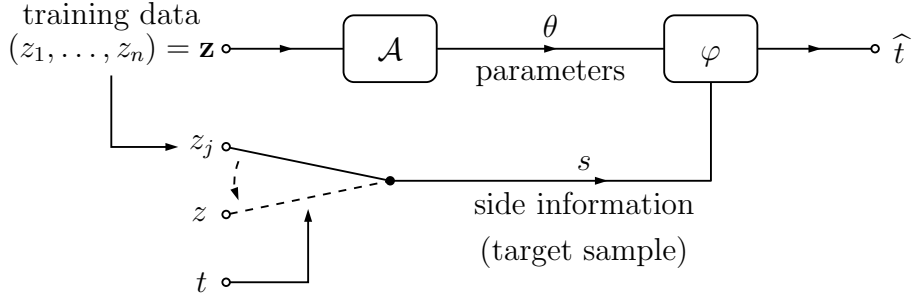

A \gls{mia} using an arbitrary strategy is illustrated in \cref{fig:ArbitraryMIA}.

% \begin{figure}
%     \centering
%     \includegraphics[width=\textwidth]{Images/MIA.PNG}
%     \caption{Scheme of a \gls{mia} using an arbitrary attack strategy. If $T=1$, the target sample is drawn from the training set used to train the target model. If $T=0$, the target sample is drawn from the data distribution.}
%     \label{fig:ArbitraryMIA}
% \end{figure}

Although, in practice, the prior distribution of the target attribute $T$ is usually unknown, we define the optimal rejection region of an idealized attacker, having access to all other involved distributions.

\begin{definition}[Most powerful test  {according to Neyman-Pearson lemma}]\label{def-decision-region}
In a membership inference setup (\cref{def:memb_inf}), define, for a threshold $0 < \gamma < \infty$, the decision region 
\begin{align}
    \DecisionSet(\gamma) 
    \triangleq \left\{(\theta,s) \in\Theta\times\mathcal{S}:{p_{\widehat{\theta}(\mathbf{Z})S|T}\big(\theta,s |1\big)}> \gamma\cdot{p_{\widehat{\theta}(\mathbf{Z})S|T}\big(\theta,s | 0\big)}\right\}.\label{eq:opt}
\end{align}
{By the Neyman-Pearson lemma~\cite{neymanpearson1933},} the most powerful test at threshold $\gamma$ is then {given} by detecting $T = 1$ for all pairs $(\theta,s) \in \DecisionSet(\gamma)$, i.e., $\varphi(\theta, s) = 1$ if and only if $(\theta, s) \in \DecisionSet(\gamma)$.
\end{definition}
In \cref{prop:1} we will provide lower bounds on the error achieved by this decision region and make the connection to the fully Bayesian case.
%%% Local Variables:
%%% mode: latex
%%% TeX-master: "main"
%%% End:

%% file: Images/mia.tikz
% \tikzstyle{every pin edge}=[<-,shorten <=1pt]
% \tikzstyle{multiple}=[line width=2pt]
% \tikzstyle{neuron}=[circle,fill=black!25,minimum size=17pt,inner sep=2pt]
% \tikzstyle{input neuron}=[neuron, fill=green!50]
% \tikzstyle{output neuron}=[neuron, fill=red!50]
% \tikzstyle{hidden neuron}=[neuron, fill=blue!50]
% \tikzstyle{box}=[rectangle, draw, rounded corners, minimum height=2em, minimum width=2.5em]
% \tikzstyle{annot} = [text width=4em, text centered]
\tikzstyle{box}=[dspsquare, rounded corners, minimum height=2em, minimum width=3em, inner sep=5pt]

\begin{tikzpicture} %, node distance=\layersep]
  \matrix (m) [row sep=8mm, column sep=15mm]
  {
    % --------------------------------------------------------------------
    % \node[dspnodeopen,dsp/label=left]  (m00) {$(z_1, \dots, z_n) = \mathbf z$};          &
    \node[dspnodeopen]                  (m00) {};       &
    \node[box]                         (m02) {$\mathcal A$}; &[15mm]
    \node[box]                  (m03) {$\varphi$};       &
    \node[dspnodeopen,dsp/label=right]                  (m04) {$\widehat t$};       \\
  
    % --------------------------------------------------------------------
    \node[dspnodeopen]                  (m20) {};          &
    \node[coordinate]                  (m21) {};          &
    \node[coordinate]                  (m22) {};          &
    \node[coordinate]                         (m23) {};          &
    \node[coordinate] (m24) {};          \\
    % --------------------------------------------------------------------
    \node[dspnodeopen,dsp/label=left]  (m40) {$z$};    &
    \node[coordinate]                  (m41) {};            &
    \node[coordinate]                  (m42) {};    &
    \node[coordinate]                  (m43) {};            &
    \node[coordinate]                  (m44) {};    \\
    % --------------------------------------------------------------------
    \node[dspnodeopen,dsp/label=left]                  (t) {$t$};            &
    \node[coordinate]                  (m51) {};            &
    \node[coordinate]                  (m52) {};    &
    \node[coordinate]                  (m53) {};            &
    \node[coordinate]                  (m54) {};    \\
    % -------------------------------------------------------------------- 
  };

  \node (Svert) at ($(m20)!0.5!(m40)$) {};
  \node[dspnodefull] (S) at (Svert -| m02) {};

  \node[anchor=east] (z) at (m00) {$(z_1, \dots, z_n) = \mathbf z$};
  \node[anchor=east] (zj) at (m20) {$z_j$};

  % Draw connections

  \draw[dspflow] (m00) -- (m02);
  \draw[dspflow] (m02) -- node [above] {$\theta$} node [below] {parameters} (m03);

  \draw[dspline] (m20) -- (S);

  \node (slii) at ($(m40)!0.1!(S)$) {};
  \node (sli) at ($(m20)!0.1!(S)$) {};
  
  \draw[dspline,dashed] (m40) -- node[name=sl] {}  (S) ;

  \draw[dspflow] (S)  -| node [pos=0.3,above] {$s$} node [pos=0.3,below,name=slabel] {side information} (m03);
  \node[align=center,anchor=north] at (slabel.south) {(target sample)};
  \draw[dspflow] (m03) -- (m04);

  \draw[dspconn] (t) -| (sl);
  \draw[dspconn] (z) |- (zj);
  
  \node[anchor=south,above=.2\baselineskip] at (z) {training data};

  \draw[dspconn,dashed] (sli) to [out=250,in=110] (slii);

  % \foreach \i [evaluate = \i as \j using int(\i+1),
  % evaluate = \i as \k using int(\i+2),] in {2,4,6}
  % {
  % \begin{scope}[start chain]
  %   \chainin (m0\i);
  %   \chainin (m0\j) [join=by dspconn];
  %   \chainin (m0\k) [join=by dspflow];
  %   \chainin (m1\k) [join=by dspconn];
  %   \chainin (m2\k) [join=by dspconn];
  % \end{scope}
  % \draw[dspconn] (m2\i) -- (m2\k);
  % }

  %   \draw[dspflow] (m28) -- (m2X);

\end{tikzpicture}

%% file: mainresults.tex
\subsection{Performance of the Bayesian Attacker}
\label{sec:Oracle}

% \begin{figure}
%     \centering
%     \includegraphics[width=\textwidth]{Images/BayesianAttacker.PNG}
%     \caption{Scheme of the Bayesian attacker. The Bayesian attacker achieves the upper bound shown in \cref{thm:neyman_pearson_optimality}, but requires estimation of the conditional distribution of the target attribute. The observations required for the attack are side-information $s$ and model parameters $\theta$.}
%     \label{fig:BayesAtt}
% \end{figure}

In this section, we establish two theorems that provide upper bounds on the success probability of an arbitrary attacker. First, consider the general case in which the target attribute $T$ is not necessarily binary, but finite. This case includes both membership and feature inference attacks. In this case the Bayes classifier is the best possible attacker\added{, which arises naturally from a maximum a posteriori optimization of the target attribute.}

\begin{theorem}[Success of the Bayesian attacker]
    \label{thm:neyman_pearson_optimality}
    Assume that $\TTT$ is a finite set and $\varphi$ is an arbitrary attack strategy.\footnote{\added{As this result provides an upper bound on the success probability, no restrictions are placed on the capabilities of the attacker.}} The Bayes success probability is upper bounded by,
    \begin{align}
        \PS(\varphi) \le \EE\bigg[\max_{t\in\TTT}p_{T|\widehat\theta(\vt Z) S}(t|\widehat \theta(\vt Z), S)\bigg] \;,
        \label{eq:thm1:1}
    \end{align}
    where the upper bound is achieved by the attack strategy,
    \begin{align}
        \varphi^{\star}(\theta, s) &= \argmax_{t \in \TTT} p_{T|\widehat\theta(\vt Z) S}(t|\theta, s)\;.
        \label{eq:thm1:2}
    \end{align}
    If the $\argmax$ in \cref{eq:thm1:2} is not unique, any $t \in \mathcal{T}$ achieving the maximum can be chosen.
\end{theorem}

\begin{proof}
    Let $\widehat T$ denote the random variable defined by $\widehat T \triangleq \varphi(\widehat\theta(\vt Z), S)$. Note that $\widehat T$ is independent from $T$ given $(\widehat\theta(\vt Z), S)$. First, the upper bound in \cref{eq:thm1:1} is shown, then it is shown that this upper bound is achieved by \cref{eq:thm1:2}. Let $\varphi$ be an arbitrary attack strategy defining pdf $p_{\widehat T|\widehat\theta(\vt Z) S}(\widehat t|\theta,s)$ for each $(\theta,s)\in\Theta\times\SSS$,
    \begin{align}
        \PS(\varphi) &= \EE\bigg[\sum_{\widehat t\in\TTT}p_{\widehat T|\widehat\theta(\vt Z) S}(\widehat t|\widehat \theta(\vt Z), S){p_{T|\widehat\theta(\vt Z) S}(\widehat t|\widehat \theta(\vt Z), S)}\bigg]
        \nonumber\\
        &\leq\EE\bigg[\max_{t'\in\TTT}p_{T|\widehat\theta(\vt Z) S}(t'|\widehat \theta(\vt Z), S)\bigg]\;.
    \end{align}
    Now, consider an attack strategy $\varphi^{\star}$, such that $\varphi^{\star}(\theta, s)$ is in
    \begin{align}
         \left\{t\in\TTT:p_{T|\widehat\theta(\vt Z) S}(t|\theta, s)=\max_{t'\in\TTT}p_{T|\widehat\theta(\vt Z) S}(t'|\theta, s)\right\}\;, \label{eq-best-phi}
    \end{align}
    for given $\theta \in \Theta$ and $s \in \SSS$. Hence,
    \begin{align}
        \PS(\varphi^{\star}) =&\EE\bigg[\max_{t'\in\TTT}p_{T|\widehat\theta(\vt Z) S}(t'|\widehat \theta(\vt Z), S)\bigg]\;.
    \end{align}
    Note that the bound is achieved as long as \cref{eq-best-phi} is satisfied.
\end{proof}

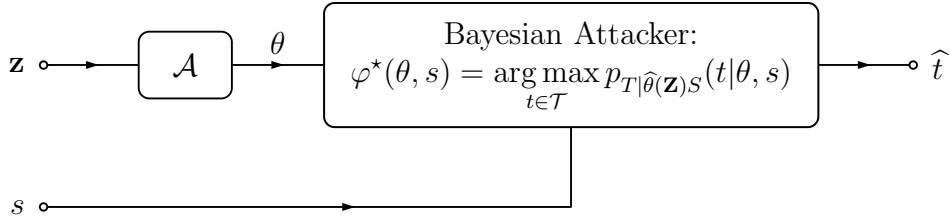
\begin{figure}
    \centering
    \input{Images/bayes_attack.tikz}
    \caption{Scheme of the Bayesian attacker. The Bayesian attacker achieves the upper bound shown in \cref{thm:neyman_pearson_optimality}, but needs to be able to evaluate the conditional distribution $p_{T|\widehat\theta(\vt Z) S}$. The observations required for the attack are the side-information $s$ and model parameters $\theta$.}
    \label{fig:BayesAtt}
\end{figure}

\added{A schema of the Bayesian attack is shown in \cref{fig:BayesAtt}.} Given white-box access to the model and its parameters, as well as side information, the attacker \cref{eq:thm1:2} has the highest probability of successfully identifying a record in the training set.  Thus, resilience against strategy \cref{eq:thm1:2} provides a strong privacy guarantee. \added{Note that, even though $S$ plays a very specific role in a \gls{mia}, it may contain additional samples, or any other kind of information, making \cref{thm:neyman_pearson_optimality} applicable to other setups.}

\Cref{thm:neyman_pearson_optimality} \added{can also be applied to the black-box case. A black-box attack is not granted access to the parameters $\theta \in \Theta$, but only to the input-output relation  $\big\{ \big(x,f_\theta(x)\big) \mid x\in \XXX \big\} $ where  $f_\theta\in \FFF$ is  the model associated to the parameters $\theta$. Thus, any black-box attack strategy $\varphi':  \FFF \times \SSS \to \TTT$ can be seen as a particular case of a white-box strategy defined as $\varphi(\theta, s) = \varphi'(f_\theta, s)$, and therefore the upper bound expressed by} \cref{thm:neyman_pearson_optimality} \added{still applies, since it is an upper bound for \emph{all} strategies.}

\added{
Similarly, when the attacker has access to only a subset of the parameters, it can be seen as a particular case of the attacker considered in \cref{thm:neyman_pearson_optimality}, and therefore the result  still applies.
}

The following proposition provides similar results for the membership inference problem.

\begin{proposition}[Decision tradeoff]
\label{prop:1}
	In a membership inference setup (\cref{def:memb_inf}), let $\DecisionSet\subseteq  \Theta\times\mathcal{S}$ be any decision set, and define 
	\begin{equation}
    \epsilon_{1}(\DecisionSet) \triangleq  \int_{\DecisionSet}p_{\widehat{\theta}(\mathbf{Z})S|T}(\theta,s | 0) \,d\theta \,ds \;,\;
    {\epsilon}_{0}(\DecisionSet^c) \triangleq  \int_{\DecisionSet^c}p_{\widehat{\theta}(\mathbf{Z})S|T}(\theta,s | 1)\,d\theta \,ds,  \label{eq:type2}
    \end{equation}
	the average Type-I (false positive) and Type-II (false negative) error probabilities, respectively. Then, 
	\begin{align}
		{\epsilon}_{0}(\DecisionSet) + {\epsilon}_{1}(\DecisionSet^c) &\geq  1-\Delta ,
	    \label{eq:prop1}
	\end{align}
	where $ \Delta \triangleq \big\Vert p_{\widehat{\theta}(\mathbf{Z})S|T=1} - p_{\widehat{\theta}(\mathbf{Z})S|T=0} \big\Vert_\mathrm{TV}\;$ and 
	$\Vert\cdot\Vert_\mathrm{TV}$ is the total variation distance~\cite{Gibbs2002choosing}. Equality is achieved by choosing  $\DecisionSet^\star \equiv \DecisionSet(1)$ according to \cref{def-decision-region}. If the hypotheses are equality distributed, then the minimum average Bayesian error satisfies 
	\begin{equation}
 	    \inf_\varphi\,  \Prob\left\{\varphi(\widehat{\theta}(\mathbf{Z}),S)\neq T \right\} =  \frac12 \left( 1 -  \Delta\right)\;.
 	    \label{eq:prop1:2}
	\end{equation}
\end{proposition}

The proof of this proposition is rather lengthy and  so is relegated to \cref{app:prop1}. Equation~\cref{eq:prop1}, similar to \cref{eq:thm1:1}, provides a lower bound for the total error of an arbitrary attacker. Equation~\cref{eq:prop1:2} provides the error of the Bayesian attacker from \cref{thm:neyman_pearson_optimality} in the case where the hypotheses are equally distributed.

\subsection{Generalization Gap and Success of the Attacker}
\label{sec:GenErr}

In this section, we explore the connection between the generalization gap and the success probability of \glspl{mia}. Large generalization gap implies poor privacy guarantees against \glspl{mia}. Moreover, depending on characteristics of the loss function, the probability of success of the attacker is lower bounded by the generalization gap:
\begin{theorem}[Bounded loss function]
  \label{thm:bounded_loss}
  If the loss is bounded by $|\ell| \le \ellmax$, then
  there is an attack strategy $\varphi$ for a \gls{mia} (\cref{def:memb_inf}) such that,
  \begin{equation*}
    \PS(\varphi) \ge \max\left\{P_{\mathrm{m}}, P_{\mathrm{m}}\left( \frac{|\EG(\Alg)|}{2 \ellmax} - 1\right) + 1\right\}\; ,
  \end{equation*}
  where $P_{\mathrm{m}}\triangleq \max_{t\in\{0,1\}} \Prob\{T=t\}$.
\end{theorem}

\begin{proof}
  Recalling \cref{def:mem_inference,def:expected_risks}, \added{and in particular $\varrho(\theta,(x,y)) = \EE[\ell(\hat Y_{\theta}(x), y)]$}, as well as the random variable $R = \varrho(\widehat\theta(\vt Z), S)$, we obtain
  \begin{align}
    \label{eq:1}
    |\EG(\Alg)| &= \left| \int r (p_{R|T}(r|0) - p_{R|T}(r|1)) \, dr \right| 
    \nonumber\\
    &\le \int |r| |p_{R|T}(r|0) - p_{R|T}(r|1)| \, dr 
    \nonumber\\
    &\le \ellmax \Vert p_{R|T}(\cdot|0) - p_{R|T}(\cdot | 1) \Vert_1 .
  \end{align}
  Assume w.l.o.g.\ that the attacker $\varphi$ satisfies the condition,
  \begin{align}
    p_{RT}(\varrho(\theta, (x, y)), \varphi(\theta, x, y)) \ge p_{RT}(\varrho(\theta, (x, y)), 1 - \varphi(\theta, x, y)) .
  \end{align}
  Thus, we obtain,
  \begin{align}
    \PS(\varphi) &= \frac{1}{2} \left(1 + \int |p_{RT}(r,0) - p_{RT}(r,1)| \, dr \right) 
    \nonumber\\
    &\ge \frac{1}{2} P_{\mathrm{m}} \Vert p_{R|T}(\cdot|0) - p_{R|T}(\cdot | 1) \Vert_1 + 1 - P_{\mathrm{m}}   
   \nonumber\\
    &\ge P_{\mathrm{m}} \left( \frac{|\EG(\Alg)|}{2 \ellmax} - 1\right) + 1. \label{eq:3}
  \end{align}
  Note that the lower bound, $P_{\mathrm{m}} \left( \frac{|\EG(\Alg)|}{2 \ellmax} - 1\right) + 1$, varies from $1-P_{\mathrm{m}}$ to $1$, as the generalization gap increases. However, an attacker with knowledge of the prior on $T$ can always have a success probability of at least $P_{\mathrm{m}}$ by guessing $\widehat{t}=\argmax_{t\in\TTT}\Prob\{T=t\}$; therefore,
  \begin{align}
      \PS(\varphi) &\ge \max\left\{P_{\mathrm{m}},P_{\mathrm{m}} \left( \frac{|\EG(\Alg)|}{2 \ellmax} - 1\right) + 1\right\} \nonumber \qedhere
  \end{align}
\end{proof}

\Cref{thm:bounded_loss} indicates that strong privacy guarantees (i.e., small success probability for any attacker), imply that the generalization gap is also small.
We remark that, on the other hand, ensuring that the generalization gap is small does not make a model robust against \glspl{mia}. We shall return to this important point in \cref{sec:good-gener-not}.

In the following, we extend the result of \cref{thm:bounded_loss} to sub-Gaussian and exponentially tail-bounded loss functions.

\begin{theorem}[Sub-Gaussian loss]
  \label{thm:MSE_gaussian}
  In a membership inference problem (\cref{def:memb_inf}), assume that $R = \varrho(\widehat\theta(\vt Z), S)$ is a sub-Gaussian random variable with variance proxy $\sigma^2_R$. For all $\Rmax \ge r_0 \triangleq \sqrt{2 \sigma_R^2 \log 2}$, there exists an attack strategy $\varphi$, such that,
  \begin{equation}
    \PS(\varphi) \ge \max\bigg\{ P_{\mathrm{m}}, P_{\mathrm{m}} \bigg( \frac{|\EG(\Alg)|}{2 \Rmax} - \frac{C(\Rmax, \sigma_R)}{1-P_{\mathrm{m}}} - 1\bigg) + 1\bigg\}\;.
    \label{eq:th:psuc}
  \end{equation}
  where $C(\Rmax, \sigma_R) \triangleq \exp\left(-\frac{\Rmax^2}{2\sigma_R^2}\right) \left(1 + \frac{\sigma_R^2}{\Rmax^2} \right)$.
\end{theorem}

\begin{proof}
 Given that $R$ is a sub-Gaussian random variable with variance proxy $\sigma_R^2$, we have $\Prob\{|R| \ge r \} \le 2 e^{-\frac{r^2}{2\sigma_R^2}}$ for all $r \ge 0$ \cite{Buldygin1980Sub}. Define the random variable $R_0$ to have the distribution function $Q_0(r) \triangleq \Prob\{R_0 \le r \} \triangleq 1 - 2 e^{-\frac{r^2}{2\sigma_R^2}}$ on its support $[r_0, \infty)$, where $r_0 = \sqrt{2 \sigma_R^2 \log 2}$, i.e., the \gls{pdf} of $R_0$ is $p_{R_0}(r) = \frac{2r}{\sigma_R^2} e^{-\frac{r^2}{2\sigma_R^2}}$. Let $Q$ be the distribution function of $|R|$. Then, using the construction in the proof of \cite[Theorem~1.104]{Klenke2013Probability}, we can write $|R| = Q^{-1} \circ Q_0(R_0)$, where $Q^{-1}$ is the left continuous inverse of $Q$, noting that $Q_0$ is continuous.  The sub-Gaussian property then implies $Q(r) = 1 - \Prob\{|R| \ge r \} \ge Q_0(r)$, which immediately yields $Q^{-1} \circ Q_0(r) \le r$.

  We thus have, for $\Rmax \ge r_0$,
  \begin{align}
    \label{eq:4}
    \int_{|r| \ge \Rmax} |r| p_R(r) dr &= \int_{Q_0(r) \ge Q(\Rmax)} \hspace{-15pt} Q^{-1}(Q_0(r)) p_{R_0}(r)dr 
    \nonumber\\
    &\le \int_{ Q_0(r) \ge Q(\Rmax)} r p_{R_0}(r)dr 
    \nonumber\\
    &\le \int_{r \ge \Rmax} r p_{R_0}(r)dr 
    \nonumber\\
    &\le 2\Rmax C(\Rmax, \sigma_R)
  \end{align}
  Following steps similar to those in \cref{eq:1},
    \begin{align}
        |\EG(\Alg)| &\le \int_{|r| \le \Rmax} |r| |p_{R|T}(r|0) - p_{R|T}(r|1)| \, dr 
        \nonumber\\
        &+ \int_{|r| > \Rmax} |r| |p_{R|T}(r|0) - p_{R|T}(r|1)| \, dr 
        \nonumber\\
        &\le \Rmax \Vert p_{R|T}(r|0) - p_{R|T}(r|1) \Vert_1 +\frac{2 \Rmax C(\Rmax, \sigma_R)}{1-P_\mathrm{m}}\;,
    \end{align}
    where the last inequality follows from \cref{eq:4}. Consequently,
    \begin{align}
      \Vert p_{R|T}(r|0) - p_{R|T}(r|1) \Vert_1 \ge \frac{|\EG(\Alg)|}{\Rmax} - \frac{2 C(\Rmax, \sigma_R)}{1-P_\mathrm{m}}.
    \end{align}
    The rest of the proof follows identically to that of \cref{thm:bounded_loss}.
\end{proof}

\begin{theorem}[Tail-bounded loss]
    \label{thm:MSE_Exp}
    In a membership inference problem (\cref{def:memb_inf}), assume that $R = \varrho(\widehat\theta(\vt Z), S)$ is such that $\Prob\{|R| \ge r\}\le 2\exp(-r/2\sigma_R^2)$ for all $r \ge 0$ with some variance proxy $\sigma_R^2 > 0$. Then, for all $\Rmax \ge r_0 \triangleq 2 \sigma_R^2 \log 2$, there is an attack strategy $\varphi$ such that,
    \cref{eq:th:psuc} holds with
    \begin{equation}
    C(\Rmax,\sigma_R) \triangleq \exp\left(-\frac{\Rmax}{2\sigma_R^2}\right) \left(1 + \frac{2\sigma_R^2}{\Rmax} \right).\label{eq-missing-eq}
    \end{equation}
\end{theorem}

The proof of this theorem is analogous to that of \cref{thm:MSE_gaussian} and will be omitted here.
% is relegated to \cref{appendix-ProofThm4}.

Note that in principle both \cref{thm:MSE_gaussian} and \cref{thm:MSE_Exp} are applicable when the loss is bounded, since all bounded random variables are sub-Gaussian and exponentially tail-bounded; nonetheless, we expect \cref{thm:bounded_loss} to provide a tighter bound in this case, as it certainly does for $\ellmax=\Rmax$.

In practice the distribution of the loss for a particular model is often unknown; however, it can be estimated and fitted to one of the cases presented in this section. Then, these results can be applied to measure the potential impact of generalization on the privacy leakage of the model. 

\subsection{Good Generalization is not Enough to Prevent Successful Attacks}
\label{sec:good-gener-not}

\emph{Generalization does not imply privacy}. The purpose of this section is to prove that in general the success rate of the attacker may not be directly proportional to the generalization gap. We show this by constructing a synthetic example of a membership inference problem, where the generalization gap can be made arbitrarily small, while $T$ can be determined with certainty by an attacker. To construct the counterexample we need to define the random variables $X$, $Y$ and a loss function $\ell$ for fixed parameters $0 < \delta < D$.
Let $p_{X}$ be an arbitrary continuous pdf\ on $\mathbb R$, e.g., $X \sim \mathcal N(0,\sigma^2)$, and define $Y = X + U$, where $U$ is independent of $X$ and uniformly distributed on $[-\frac{\smallparameter}{2}, \frac{\smallparameter}{2}]$.
Given the training set $\TrainSet$ and an input $x$, the learned decision function $f({}\cdot{};x)$ either outputs the correct label $y$, if $(x,y) \in \TrainSet$, and otherwise $f({}\cdot{};x) = x + D + U'$, where $U'$ is an i.i.d.\ copy of $U$, i.e., uniformly distributed on $[-\frac{\smallparameter}{2}, \frac{\smallparameter}{2}]$. With Euclidean distance loss $\ell(y, y') = |y-y'|$, these definitions immediately yield
$\Prob\{R=0|T=1\} = 1$ and the conditional pdf
\begin{equation}
  \label{eq:pRT}
  p_{R|T}(r|0) = \frac{1}{\smallparameter}\Lambda((r-D)/\smallparameter) .
\end{equation}
where $\Lambda(r) \triangleq \max(1-|r|,0)$ is the triangle distribution. The parameters $0 < \smallparameter < D$ can be chosen arbitrarily. Clearly then an attacker can simply check whether $R=0$ to determine $T$ with probability one. On the other hand, from~\cref{eq:pRT}, it is easily verified that,
\begin{align}
    |\EG(\Alg)| &= \left| \EE[R|T=0] - \EE[R|T=1] \right| = D .
\end{align}
Thus, by varying the parameter $D$, we can make the generalization gap arbitrarily small, while the attacker maintains perfect success. Therefore, good generalization does not prevent the attacker from easily determining which samples were part of the training set. Remark that as NNs are universal approximators, any (reasonable) function, including the decision rule in this example, can be approximated to arbitrary degree by a NN; therefore, this behavior could be seen in practice.

\subsection{On the Amount of Missing Information in Inference Attacks and Generalization} \label{sec:MutualInfo}
We aim at  investigating the following simple but fundamental questions, from the perspective of information theory:
\begin{itemize}
    \itemsep1mm
    \item \emph{How much information do the model parameters $\widehat\theta(\TrainSet)$ store about the training set $\TrainSet$? How is this information related to the generalization gap?}
    \item  \emph{How much information about the unknown (sensitive) attribute $T$ is contained in the model parameters $\widehat\theta(\TrainSet)$ and the side information $S$? And how much information is needed for the inference of $T$? }
    \item  \emph{How do the above information quantities relate or bound to each other? }
\end{itemize}
From the point of view of information theory these questions make sense only if we consider $\widehat\theta(\TrainSet)$ and $T$ as random variables, that is, attribute probabilities to the target attribute and model parameters, which is perfectly consistent with the investigated framework in this paper.

\added{To state the following \lcnamecref{the4}, we need the \emph{Fenchel-Legendre dual function}~\cite{10.5555/2028633} $g^\star\colon \mathbb R \to \mathbb R$ of a function $g \colon \mathbb R \to \mathbb R$, which is defined as
  $g^\star(t) \triangleq \sup \{\lambda \cdot t - g(\lambda) : \lambda \in \mathbb R\}$.
  We will also use the log-moment-generating function $\psi_W\colon \mathbb R \to \mathbb R$ of a random variable $W$, defined as $\psi_W(\lambda) \triangleq \log \EE[e^{\lambda W}]$. More information on these quantities and their properties are given in the discussion of the Cram\'er-Chernoff Method in~\cref{app:cramer-chernoff}.}

\begin{theorem}[Mutual information]
\label{the4}
Let $\widehat T \triangleq \varphi(\widehat\theta(\vt Z), S)$ be the (random) prediction of any attacker $\varphi$ (\cref{def:success_prob}). Then, 
\begin{align}
I \big(T; \widehat{\theta}(\mathbf{Z}) \big| S \big)  &\geq 
    d_{\text{KL}}\Big( \PS(\varphi) \,  \Big\| \,  \max_{t\in\mathcal{T}}\, p_{T} (t)   \Big),\label{eq-inequality-missing1}
\end{align}
where $d_{\text{KL}}( p \|q )$ denotes the KL divergence between Bernoulli random variables with probabilities $(p,q)$. Moreover, for $\epsilon \ge 0$, the generalization gap $\EG$ at $\vt Z$ satisfies 
\begin{align}
  \Prob \big( \GG(\Alg) \ge \epsilon  \big) &\leq \frac{I(\mathbf{Z}; \widehat{\theta}(\mathbf{Z}) )+1}{{n K(\epsilon) }}
  \label{eq-inequality-missing2} , 
\end{align}
where 
\begin{equation}
K(\epsilon)\triangleq  \essinf_{\theta \sim P_{\widehat\theta(\mathbf{Z})}}  \psi_{ { \mathbb{E}[\varrho(\widehat\theta,(X,Y))] - \varrho(\theta,(X,Y))}}^{\ast}(\epsilon)
\end{equation}
is an essential infimum w.r.t.\ $\theta \sim P_{\widehat\theta(\mathbf{Z})}$ of the Fenchel-Legendre dual function $\psi^\star$ of {the log-moment-generating function of} $ \mathbb{E}[\varrho(\theta,(X,Y))] - \varrho(\theta,(X,Y)) $. Furthermore, 
\begin{equation}
  I(T; \widehat{\theta}(\mathbf{Z}) | S) =   I(S; \widehat{\theta}(\mathbf{Z}) | T)  - I( S;\widehat{\theta}(\mathbf{Z}) ) \leq   I(\mathbf{Z}; \widehat{\theta}(\mathbf{Z}) )  - I( S;\widehat{\theta}(\mathbf{Z}) ).  \label{eq-inequality-missing3} 
\end{equation}
\end{theorem}

\added{\Cref{the4} is proved in} \cref{appendix-Theorem-info}.

The mutual information expressions in \cref{eq-inequality-missing1} and \cref{eq-inequality-missing2} are related by the inequality \cref{eq-inequality-missing3},
where $I(\mathbf{Z}; \widehat{\theta}(\mathbf{Z}) )$ represents the average amount of information about the random training set $\mathbf{Z}$ retained in the model parameters $\widehat{\theta}(\mathbf{Z})$; and $I(S;\widehat{\theta}(\mathbf{Z}) )$ indicates the amount of information already  contained in the side information $S$ before observing the parameters $\widehat{\theta}(\mathbf{Z})$. 

From \cref{eq-inequality-missing3} it is clear that by controlling the average number of bits of information about the training set $\mathbf{Z}$ that the model parameters $\widehat{\theta}(\TrainSet)$ store, i.e., $I(\mathbf{Z}; \widehat{\theta}(\mathbf{Z}) )  \leq r$, it is possible to control both the generalization gap in \cref{eq-inequality-missing2} and the accuracy of any possible attacker in \cref{eq-inequality-missing1}. Nevertheless, a more effective defense strategy may aim directly at reducing the mutual information $I(T; \widehat{\theta}(\mathbf{Z}) | S)$, which is expected to have less severe impact on the performance of the trained model, i.e.,  the expected risk $\EE\big[ \ell(\hat Y_{\widehat\theta(\mathbf{Z})}(X), Y)\big]$.
As~\cref{eq-inequality-missing1} indicates, the performance of any attacker must be close to a random guess if the mutual information $I(T; \widehat{\theta}(\mathbf{Z}) | S)$ is suitably small. {This equation can be numerically computed to obtain an upper bound on $\PS(\varphi )$}.

% Moreover, an explicit bound can be given: 
% \begin{equation}
%     \PS(\varphi )  \leq  \min \left \{ 
%     \sqrt{ \frac12 I(T; \widehat{\theta}(\mathbf{Z}) | S) } + \max_{t\in\mathcal{T}}\, p_{T} (t) , 
%                               \frac{I(T; \widehat{\theta}(\mathbf{Z}) | S) +1 }{ -\log_2  \max_{t\in\mathcal{T}} \, p_{T} (t) }\right\}. \label{eq-explicit-expression}
%                                 \nonumber
% \end{equation}
The generalization gap bound in \cref{eq-inequality-missing2} is subtly different from most PAC-Bayes scenarios of learning. In the present case, we are bounding the joint probability over both the training data $\mathbf{Z}$ and the randomness involved in the learning algorithm, which is within the spirit of the work by \cite{pmlr-v83-bassily18a}. But due to the term $K(\epsilon)$, the bound presented in \cref{eq-inequality-missing2} is tighter. 

Assuming that the loss is sub-Gaussian or bounded, it is not difficult to provide a lower bound for $K(\epsilon)$ that is independent of the underlying data distribution.

%%% Local Variables:
%%% mode: latex
%%% TeX-master: "main"
%%% End:

%% file: Images/bayes_attack.tikz
% \tikzstyle{every pin edge}=[<-,shorten <=1pt]
% \tikzstyle{multiple}=[line width=2pt]
% \tikzstyle{neuron}=[circle,fill=black!25,minimum size=17pt,inner sep=2pt]
% \tikzstyle{input neuron}=[neuron, fill=green!50]
% \tikzstyle{output neuron}=[neuron, fill=red!50]
% \tikzstyle{hidden neuron}=[neuron, fill=blue!50]
% \tikzstyle{box}=[rectangle, draw, rounded corners, minimum height=2em, minimum width=2.5em]
% \tikzstyle{annot} = [text width=4em, text centered]
\tikzstyle{box}=[dspsquare, rounded corners, minimum height=2em, minimum width=3em, inner sep=5pt]
\tikzstyle{box2}=[rectangle, draw, rounded corners, minimum height=4em, minimum width=3em, inner sep=5pt,line width=0.75pt]

\begin{tikzpicture} %, node distance=\layersep]
  \matrix (m) [row sep=10mm, column sep=12mm]
  {
    % --------------------------------------------------------------------
    \node[dspnodeopen,dsp/label=left]  (z) {$\mathbf z$};          &
    \node[box]                         (m02) {$\mathcal A$}; &
    \node[box2]                  (m03) {
      $\begin{array}{c}
         \text{Bayesian Attacker:} \\
         \displaystyle\varphi^{\star}(\theta, s) = \argmax_{t \in \TTT} p_{T|\widehat\theta(\vt Z) S}(t|\theta, s)
      \end{array}$
      };       &
    \node[dspnodeopen,dsp/label=right]                  (m04) {$\widehat t$};       \\
  
    % --------------------------------------------------------------------
    \node[dspnodeopen,dsp/label=left]                  (s) {$s$};          &
    \node[coordinate]                  (m21) {};          &
    \node[coordinate]                  (m22) {};          &
    \node[coordinate]                         (m23) {};          \\
    % --------------------------------------------------------------------
  };

  % Draw connections

  \draw[dspflow] (z) -- (m02);
  \draw[dspflow] (m02) -- node [above] {$\theta$} (m03);

  \draw[dspflow] (s)  -|  (m03);
  \draw[dspflow] (m03) -- (m04);

  % \foreach \i [evaluate = \i as \j using int(\i+1),
  % evaluate = \i as \k using int(\i+2),] in {2,4,6}
  % {
  % \begin{scope}[start chain]
  %   \chainin (m0\i);
  %   \chainin (m0\j) [join=by dspconn];
  %   \chainin (m0\k) [join=by dspflow];
  %   \chainin (m1\k) [join=by dspconn];
  %   \chainin (m2\k) [join=by dspconn];
  % \end{scope}
  % \draw[dspconn] (m2\i) -- (m2\k);
  % }

  %   \draw[dspflow] (m28) -- (m2X);

\end{tikzpicture}

%% file: experiments.tex
\subsection{Linear Regression on (Synthetic) Gaussian Data}
\label{sec:gaussians}

The following example allows us to illustrate how the theoretical results from the previous section might be used to assess the privacy guarantees of a specific model. We implement the Bayesian attacker from \cref{thm:neyman_pearson_optimality} and estimate its success probability to monitor the privacy leakage of the model as a function of the number of training samples. Second, since the loss is tail-bounded exponentially, we use \cref{thm:MSE_Exp} to derive lower bounds on the success probability of the attacker. Lastly, we utilize \cref{eq-inequality-missing1} from \cref{the4} to upper bound the success probability of the Bayesian attacker.

For $i\in[n]$, let $x_i$ be a fixed vector on $\RR^d$ and for a fixed vector $\beta \in \RR^d$, let $Y_i = \beta^T x_i + \Noise_i$ with $\EE[\Noise_i] = 0$ and $\EE[\Noise_i^2] = \sigma^2 < \infty$ for $i\in[n]$. The training set is $\TrainSet = \{y_1,\ldots,y_n\}$, a realization of $Y_i$ for each $i\in[n]$. The function space $\FFF$ consists of linear regression functions $f_\theta(x_i) = \theta^T x_i$ for $\theta \in \RR^d$ and the deterministic algorithm $\Alg$ minimizes squared error on the training set and thus yields\footnote{Let $\vt x$ be the $[d \times n]$ matrix $\vt x = (x_1, x_2, \dots, x_n)$. Similarly, $\vt y = (y_1, y_2, \dots, y_n)$ and $\vt W = (W_1, W_2, \dots, W_n)$ are $[1 \times n]$ vectors.} $\widehat\theta(\vt y) = (\vt x \vt x^T)^{-1} \vt x \vt y^T$ and the associated decision function
$f_{\widehat\theta(\vt y)}(x_i) = \vt y \vt x^T (\vt x \vt x^T)^{-1}x_i$. Using squared error loss, $\ell(y, y^\prime) = (y-y^\prime)^2$, we obtain the generalization gap,
\begin{equation}
  {  \EG(\Alg) }= \frac{2d}{n} \sigma^2 \;,
    \label{eq:genErrorGaus1}
\end{equation}
A derivation of this formula is presented in \cref{appendix-GaussianExample}. Assuming the noise $\Noise$ to be Gaussian, the scalar response $\vt Y =\beta^T \vt x + \vt \Noise$ then also follows a Gaussian distribution, with $\vt \Noise$ a row vector of i.i.d.\ components. Similarly, the model parameters $\widehat\theta(\vt Y)$ are normally distributed. Now choose a test sample $S_\RDI=T(Y_\RDI)+(1-T)(Y'_\RDI)$, where $\RDI$ is an index in $[n]$, $Y_\RDI$ is the $\RDI-\mathrm{th}$ component of the (random) training set and $Y'_\RDI$ is drawn independently of the training set. Assuming a Bernoulli $1/2$ prior on the hypothesis $T$, the success probability of the Bayesian attacker $\varphi^{\star}$ is given by
\begin{align}
    \PS(\varphi^{\star}) = 1-\frac12\left[\epsilon_{0}\big(\DecisionSet(1)\big) +  \epsilon_{1}\big(\DecisionSet(1)^c\big)\right]\;,
    \label{eq:SuccProbGaussian1}
\end{align}
with the Type-I and Type-II errors defined by~\cref{eq:type2}, and the optimal decision region $\DecisionSet(1)$ defined by~\cref{eq:opt}. With posteriors defined by,
\begin{align}
    p_{S_\RDI\RDI\widehat\theta|T}(s,j,\theta|0) &=  \frac{1}{n}Q(\theta) p_{Y_j}(s)\; ,
    \\
    p_{S_\RDI\RDI\widehat\theta|T}(s,j,\theta|1) &=  \frac{1}{n}Q_{j}(\theta|s) p_{Y_j}(s)\;.
\end{align}
The index $j$ indicates the feature vector $x_j$ from which the test sample $s$ is generated. $Q(\theta)$ is the distribution of the model parameters conditioned to $T=0$. It is independent of the test sample $s$ and of the index $j$. $Q_j(\theta|s)$ is the distribution of the model parameters conditioned to $T=1$. Since, under this hypothesis, the attacker assumes $s$ is one of the samples in the training set, this conditional distribution depends on the test sample $s$ and its corresponding index $j$. The distribution of the test sample $p_{Y_j}$ is defined by $p_{Y_j}({}\cdot{})\triangleq\NNN({}\cdot{};\beta^T x_j,\sigma^2)$. $Q({}\cdot{})$ and $Q_{j}({}\cdot{}|s)$ are defined by $Q({}\cdot{})\triangleq\NNN({}\cdot{};\beta,\sigma^2 \ol x^{-1})$ and $Q_{j}({}\cdot{}|s)\triangleq\NNN\big({}\cdot{}; \beta + \ol x^{-1}x_j(s - x_j^T \beta),\sigma^2\ol x^{-1}(\mathbb{I}^{d\times d} - x_j x_j^T \ol x^{-1})\big)$, respectively, where $\ol x \triangleq \vt x \vt x^T$. These distributions are derived in \cref{appendix-GaussianExample}.

The success probability of the Bayesian attack strategy in \cref{thm:neyman_pearson_optimality} is given by~\cref{eq:SuccProbGaussian1}. In our experiments we perform a Monte Carlo estimation of the integrals in~\cref{eq:type2}, by randomly drawing $T$, $s$ and $\theta$. The posterior distributions can be computed in closed form with the above definitions. Since the loss is exponentially tail-bounded, we can apply \cref{thm:MSE_Exp} to obtain the lower bound
\begin{equation}
    \PS(\varphi^{\star}) \ge 
    \frac{1}{2} + \frac{d}{2n }\frac{\sigma^2}{\Rmax} - C(R_{\mathrm{max}}, \sigma) ,
    \label{eq:SuccProbGausLB1}
\end{equation}
where we used \cref{eq:genErrorGaus1} and $C(R_{\mathrm{max}}, \sigma)$ is defined in expression \cref{eq-missing-eq}. $\Rmax$ can be chosen to maximize the upper bound in this expression. In our experiments, we choose the optimal $\Rmax$ using the golden section search algorithm. Furthermore, from \cref{eq-inequality-missing1} we have, 
\begin{align}
I \big(S_\RDI; \widehat{\theta}(\mathbf{Y}) \big| T \big)  &\geq 
    d_{\text{KL}}\Big( \PS(\varphi) \,  \Big\| \,  \max_{t\in\mathcal{T}}\, p_{T} (t)   \Big)\;.
    \label{eq:GaussUp}
\end{align}
Note that $I \big(S_\RDI; \widehat{\theta}(\mathbf{Y}) \big| T \big)\geq I \big(T; \widehat{\theta}(\mathbf{Y}) \big| S_\RDI \big)$. The mutual information between the testing sample and the model parameters given the sensitive attribute, $I \big(S_\RDI; \widehat{\theta}(\mathbf{Y}) \big| T \big)$, can be explicitly computed in this setup; the details of this computation are relegated to \cref{appendix-GaussianExample}. Fixing the prior on the hypothesis $T$ to a Bernoulli $1/2$, we can utilize \cref{eq:GaussUp} to find an upper bound on the success probability of the Bayesian attacker. This is done by searching for the success rate $\PS(\varphi)$ that makes the l.h.s.\ of \cref{eq:GaussUp} equal to its r.h.s.\ Namely, the golden section search algorithm is used to minimize the square distance between the mutual information and the KL-divergence with respect to $\PS{(\varphi)}$.

\Cref{alg:exp1} details our simulations to estimate the success rate of the Bayesian attacker. It returns `$1$' when the attacker successfully predicts whether the test sample $s$ was part of the training set or not, and `$0$' otherwise. In our experiments we vary $n$ to study how the generalization gap and success rate of the attacker evolve as a function of the number of training samples. The dimension of the feature space is fixed to $d=20$. For each value of $n$, we fix $\vt x$ and we repeat ($10$k times) \cref{alg:exp1}  to estimate the success rate of the attacker. The feature vectors $\vt x$ are generated i.i.d.\ and then fixed for each value of $n$. Additionally, for $n$, we compute the generalization gap \cref{eq:genErrorGaus1}, which is used to compute the lower bound \cref{eq:SuccProbGausLB1}. We also compute the Mutual Information in the l.h.s.\ of \cref{eq:GaussUp}, which is used to compute the upper bound on the success probability of the attacker.
\begin{figure}
  \centering
    \begin{algorithm}[H]
      \caption{Experiment 1}
      \label{alg:exp1}
      \footnotesize
      \begin{algorithmic}[1]
        \STATE {\bfseries Input:} feature vectors $\vt x$, training set size $n$  
        \STATE Draw $t$ uniform in $\{0,1\}$
        \STATE Draw $j$ uniform in $[n]$
        \STATE $\vt y \longleftarrow \beta^T \vt x + \vt \Noise$
        \IF{$t$}
        \STATE $s\longleftarrow y_j$
        \ELSE
        \STATE $s\longleftarrow \beta^T x_j + \Noise$
        \ENDIF
        \STATE $\theta \longleftarrow (\vt x \vt x^T)^{-1} \vt x \vt y^T$
        \STATE {\bfseries return} $ p_{S_\RDI\RDI\widehat\theta|T}(s,j,\theta|1)> p_{S_\RDI\RDI\widehat\theta|T}(s,j,\theta|0)$  {\bfseries   XNOR } $t$
      \end{algorithmic}
    \end{algorithm}
\end{figure}

\begin{figure}
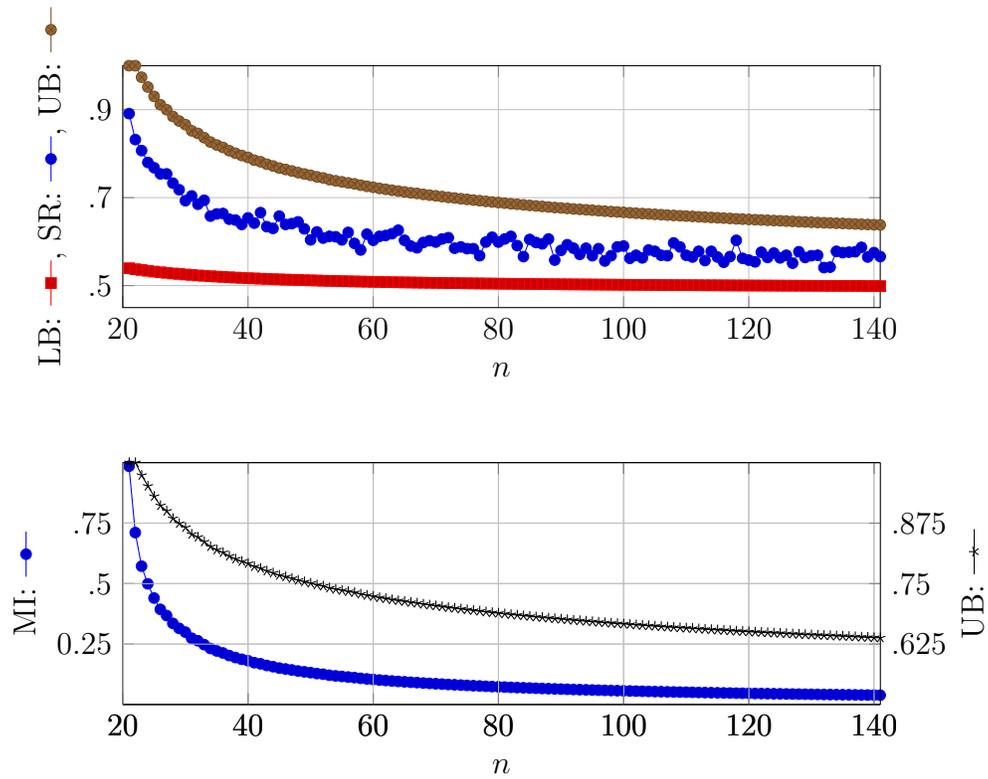

  \centering
  % \subfigure[{Success Rate (SR), Lower Bound (LB), and upper bound (UB) depend on number of training samples $n$.}]{
  %   \includestandalone{plots/SRGauss}
  %   \label{fig:Gaus:2}
  %   % }
  % }\qquad\qquad
  % \subfigure[{Mutual information (MI), Upper Bound (UB; axis labels on the right) depend on the number of training samples $n$.}]{
  %   \includestandalone{plots/MIGauss}
  %   \label{fig:Gaus:3}
  % }
  \includestandalone{plots/Gauss}
  \caption{\added{Dependence of success rate of the Bayesian attacker, generalization gap, and mutual information on the number of training samples $n$, using Gaussian data. \textbf{Top:} Success Rate (SR), Lower Bound (LB), and Upper Bound (UB). \textbf{Bottom:} Mutual Information (MI), Upper Bound (UB; axis labels on the right).}}
  \label{fig:gaussian_exp}
\end{figure}

\Cref{fig:gaussian_exp} (\textbf{Top}) shows the success rate (SR) of the Bayesian attacker as a function of the number of samples in the training set $n$. Along with it is the lower bound (LB) provided by \cref{thm:MSE_Exp} and the upper bound (UP) provided by equation \cref{eq:GaussUp}. The lower bound predicts the behavior of the SR as a function of the generalization gap. For large $n$ (small generalization gap), the success rate and its lower bound approach $0.5$, the success rate of an attacker that only uses knowledge on the prior of $T$. While the lower bound seems loose in this setting, it is worth noting that we compare with the best possible strategy. Nonetheless, this example shows that the bounds are not vacuous and they may serve as a framework for understanding the connection between information leakage and generalization in \gls{ml}. On the other hand, the upper bound provides a strong privacy guarantee. In cases where the success rate of the Bayesian attacker cannot be explicitly computed, its upper bound is the best privacy guarantee that can be provided. Additionally, \cref{fig:gaussian_exp} (\textbf{Bottom}) shows the mutual information (l.h.s.\ of \cref{eq:GaussUp}) that is used to compute the upper bound. 

\subsection{Examples on DNNs}
\label{sec:DNNs}

\begin{table*}[t]
    \centering
    \resizebox{\textwidth}{!}{
    \begin{tabular}{c|c|c|c|c|c}
         & Likelihood & Loss & Modified Entropy \cite{Song2021} & \cite{DBLP:journals/corr/abs-1709-01604} & \cite{DBLP:journals/corr/ShokriSS16} \\
         \hline
         Attack complexity & One query & One query & One query & One query & Thousands of queries and \\
         & & & & & train shadow models \\
         \hline
         Required Knowledge & Soft Probabilities & Loss value & Soft Probabilities & Training Loss & Additional Samples \\
         \hline
         PPV MNIST & $0.444\pm0.000$ & $0.446\pm0.000$ & $0.444\pm0.000$ & 0.505 & 0.517 \\
         PPV CIFAR-10 & $0.446\pm0.001$ & $0.451\pm0.001$ & $0.449\pm0.001$ & 0.694 & 0.72 \\
         PPV Fashion-MNIST & $0.445\pm0.000$ & $0.447\pm0.001$ & $0.446\pm0.001$ & -- & -- \\
         \hline
         Recall & $>0.99$ & $>0.99$ & $>0.99$ & $>0.99$ & $>0.99$
    \end{tabular}
    }
    \caption{Comparison of the likelihood attack to previous black-box \glspl{mia} from the literature. Precision (PPV; Positive Predictive Value) and recall are reported for CIFAR10, MNIST and Fashion-MNIST.}
    \label{tab:1}
\end{table*}
We train \glspl{dnn} on various datasets to study the interplay between generalization gap and the success rate of \added{three different black-box \gls{mia} strategies}. We compare the success rate of the different attack strategies to the lower bound provided by \cref{thm:bounded_loss}, to assess the quality of the bound. Our datasets for these experiments are: CIFAR10~\cite{Krizhevsky09learningmultiple}, MNIST~\cite{lecun2010mnist}, MNIST fashion~\cite{DBLP:journals/corr/abs-1708-07747}. Details about datasets, the target model and the experiments are given in \cref{app:details:DNNs}.

The loss function used for training and for computing the generalization gap is the Mean Squared Error (MSE) loss between the one-hot encoded labels and the soft probabilities output by the network. Note that this loss function is bounded by $2$. While cross-Entropy is a more common choice for loss function, it is not bounded. On the other hand MSE has a negligible effect on performance and allows us to apply \cref{thm:bounded_loss} to lower bound the success probability of the Bayesian attacker. However, in this setup it results impossible to estimate the success probability of the Bayesian attacker, due to the high number of model parameters. To circumvent this limitation and assess the quality of the bound provided by \cref{thm:bounded_loss}, we implement the likelihood attack, detailed in \cref{alg:attak1}, \added{the loss value attack, and the modified entropy attack from \cite{Song2021}}, and compare their success rate to the bound. 

\added{The \textbf{likelihood} attack exploits the level of confidence of a trained model in its prediction, based on the assumption that the model will make more confident predictions on samples that were part of its training set. This attack returns $\widehat t = 1$ if the score function
$\max_{i\in|\YYY|} f^i_{\theta}(x)$ is higher than some given threshold.}

\added{The \textbf{loss} value attack works by assuming that the loss value will be lower for samples present in the training set, since the training algorithm minimizes the loss on these samples. This attack compares the score
  $\ell\big( f_{\theta}(x),y\big)$ to some threshold.
}

\added{On the other hand, the \textbf{modified entropy} attack simultaneously considers the correctness and the entropy of the prediction. We refer the reader to \cite{Song2021} for a more detailed description. The score used by this attack is given by}
\begin{align}
  -\left(1-f_{\theta}^{y}(x)\right)\log\left(f_{\theta}^{y}(x)\right)-\sum_{i\neq y}f_{\theta}^{i}(x)\log\left(1-f_{\theta}^{i}(x)\right).
  \label{eq:score:mentr}
\end{align}

\begin{figure}
  \centering
    \begin{algorithm}[H]
      \caption{Likelihood Attack}
      \label{alg:attak1}
      \footnotesize
      \begin{algorithmic}[1]
        \STATE {\bfseries Input:} Target model $\mathrm{NN}$, threshold $h$  
        \STATE Draw $t$ uniform in $\{0,1\}$
        \IF{$t$}
        \STATE Draw $s$ uniform from the training set.
        \ELSE
        \STATE Draw $s$ uniform from the test set.
        \ENDIF
        \STATE $\mathrm{likelihood}\longleftarrow\mathbf{max}(\mathrm{NN}(s))$
        \STATE {\bfseries return} $\mathrm{likelihood}>h$  {\bfseries   XNOR } $t$
      \end{algorithmic}
    \end{algorithm}    
\end{figure}

\begin{figure}
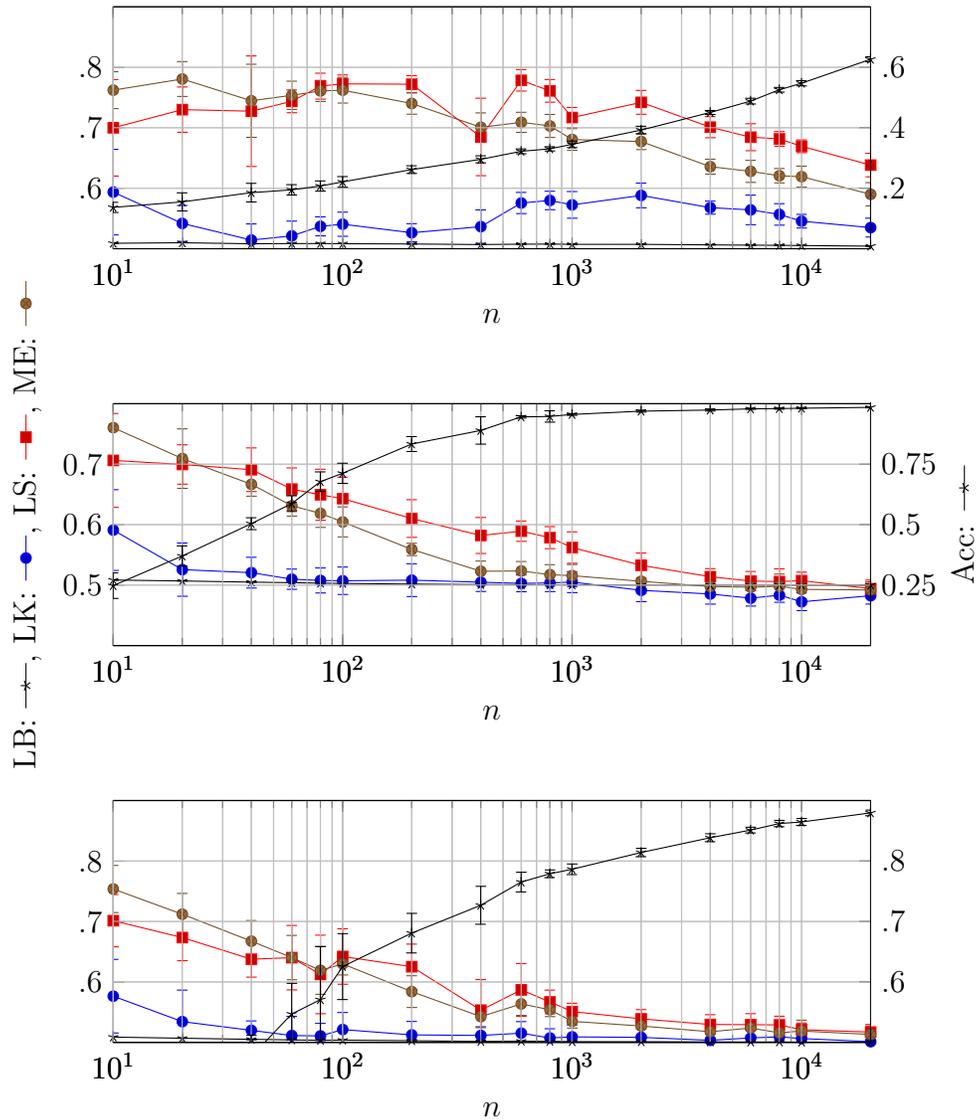

    \centering
    \includestandalone{plots/attacks}
    \caption{\added{Success rate of the likelihood attack (LK), loss value attack (LS), modified entropy attack (ME),  lower bound (LB) and accuracy (Acc; axis labels on the right) depend on the number of training samples $n$. \textbf{Top:} CIFAR10; \textbf{Middle:} MNIST; \textbf{Bottom:} FashionMNIST.}}
    \label{fig:dnns_exp}
\end{figure}

The training set is chosen uniformly at random from a pool of possible training samples. We vary the size $n$ of the training set and observe how this affects the success rate of attacks, the generalization gap and consequently the lower bound derived from \cref{thm:bounded_loss}. \added{For $n$, the number of samples in the training set, the success rate of the likelihood attack (LK), the loss value attack (LS), the modified entropy attack (ME), the lower bound (LB) provided by \cref{thm:bounded_loss}, as well as the accuracy on the test set (Acc) are obtained empirically in $100$ runs. The results for CIFAR10, MNIST and FashionMNIST are reported in \Cref{fig:dnns_exp}.} The lower bound predicts the behavior of the success rate of the likelihood attack as a function of the generalization gap; both approach $0.5$ as the generalization gap vanishes.
Note that it is possible for the success rate of the likelihood attack to go below the lower bound of the Bayesian attack. For some large $n$ values of MNIST the average success rate of the attacker goes below $0.5$. In this region the attacker cannot do better than a random guess and sometimes its success rate goes below $0.5$, which implies the model can be more confident in samples outside the training set. This is an artifact of the random sampling of the training set and the training of the model.

\Cref{tab:1} compares the strategies considered in our experiments to other previous \gls{mia} strategies found in the literature. The three strategies here considered do not require access to the model parameters or additional samples, and they only need to query the model once, while the other strategies~\cite{DBLP:journals/corr/abs-1709-01604,DBLP:journals/corr/ShokriSS16} require extra information or significantly more computing power. The attack is performed against target models with a training set of $8000$ samples, to match the setup used in \cite{DBLP:journals/corr/abs-1709-01604,DBLP:journals/corr/ShokriSS16}; however, the architectures of the target models, as well as the (random) selection of training samples differ in all three setups.

\subsection{Attribute Inference on PenDigits}
\label{sec:pendig}

To demonstrate the risk of information leakage from \gls{ml} models, we consider attribute inference attacks against a model that classifies hand-written digits. We consider the PenDigits dataset~\cite{Dua:2019}, as it contains identity information about the writers, which we use as the sensitive attribute.  The target model is a fully-connected network trained to classify hand-written digits. Details about the model and its training are provided, along with information about the dataset and its pre-processing, in \cref{app:details:PenDig}. When performing \glspl{mia}, we utilize MSE, which is bounded, as the loss for training. This allows us to apply \cref{thm:bounded_loss}.

Next, we discuss the attack strategies considered against the model. Since $\TTT$ is finite, our attack strategy consists on testing every possible value of $T$ and choosing the most likely value according to some criteria. The \textit{Gradient} and \textit{Loss} strategies are inspired by similar strategies from the membership inference literature \cite{Nasr_2019,DBLP:journals/corr/ShokriSS16}.

\textbf{Likelihood:} The intuition behind this attack is that a model is more confident on samples that were part of its training. Therefore, by choosing the correct value $t$, the model will maximize its output for the predicted label. Note that this criteria does not care about the model making the right prediction. The side information given to the attacker are the non-sensitive attributes, $s=v$. This strategy chooses the sensitive attribute that outputs the highest score, i.e.,
\begin{align}
    \varphi(v,\theta) = \argmax_{t\in\TTT} \left[\max_{i\in|\YYY|} f^i_{\theta}\big((v,t)\big)\right]\;,
\end{align}
where $f^i_\theta$ is the $i$-th component of the output of the model parameterized by $\theta$.

\begin{figure}
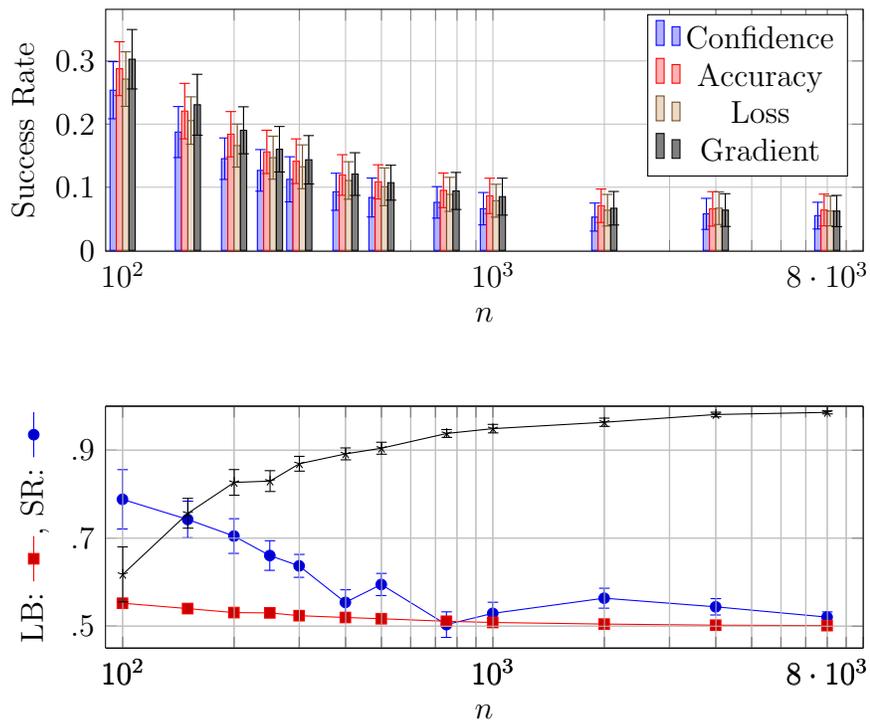

  \centering
  \includestandalone{plots/AttInf}
  \caption{\added{Attribute and Membership Inference Attacks on PenDigits for different sizes $n$ of training sets.
      \textbf{Top:} Success Rate of different attribute inference attack strategies;
      \textbf{Bottom:} Success Rate of the likelihood attack (SR), Lower Bound (LB) and Accuracy (Acc; axis labels on the right).
    }}
  \label{fig:AttInf}%
\end{figure}

\textbf{Accuracy:} In contrast to the previous one, this strategy chooses the sensitive attribute that produces the \emph{right} prediction with the highest score. This is the closest to the strategy proposed by \cite{184489}. The side information given to the attacker are the non-sensitive attributes and the label, $s=(v,y)$. Define set $\widehat{X}_{y\theta}\triangleq\{x\in\XXX:\, \argmax (f_\theta(x)) = y\}$, then,
\begin{align}
    \varphi(v,y,\theta) = \argmax_{t\in\TTT:\,x\in\widehat{X}_{y\theta}} \left[\max_{i\in|\YYY|} f^i_{\theta}\big((v,t)\big)\right]\;.
\end{align}
\textbf{Loss:} This attack, based on the value of the loss, tries to minimize the loss function over samples present in the model's training set; while the next attack uses the norm of its gradient with respect to the model parameters. The side information given to the attacker is the non-sensitive attributes and the label: $s=(v,y)$. This strategy chooses the sensitive attribute that minimizes the loss, i.e.,
\begin{align}
    \varphi(v,y,\theta) = \argmin_{t\in\TTT} \ell\left( f_{\theta}\big((v,t)\big),y\right)\;.
\end{align}
\textbf{Gradient:} Near a minimum, the norm of the gradient of the loss function with respect to its model parameters should approach $0$; the attacker exploits this knowledge for the present attack strategy. While the previous attacks only make use of the output of the model or the value of its loss, the present attack makes explicit use of its parameters. The side information given to the attacker are the non-sensitive attributes and the label, $s=(v,y)$. This strategy chooses the sensitive attribute that minimizes the gradient norm, i.e.,
\begin{align}
    \varphi(v,y,\theta) = \argmin_{t\in\TTT} \Vert\nabla_{\theta}\ell\left( f_{\theta}((v,t)),y\right)\Vert^2_2\;.
\end{align}
In our experiments we perform attribute inference attacks using each of these strategies as we vary $n$. A detailed description of the experimental procedure is given in \cref{app:details:PenDig}. The success rates for each strategy are computed and reported in \Cref{fig:AttInf} (\textbf{Top}). In this setup, a random guess would amount to a success rate of approximately $2.3\%$. For a small training set ($100$ samples), the attacker has a gain of $25\%$ over a random guess. This decreases significantly with the size of the training set; however, even for a large training set, the attacker still has twice as much accuracy as a random guess.

Additionally, we perform \glspl{mia} against the same models. The attack strategy utilized is the Likelihood attack, given in \cref{alg:attak1}. The success rate of the attacker, lower bound on the Bayesian attacker and accuracy of the model are presented in \cref{fig:AttInf} (\textbf{Bottom}). We can observe that there is a significant leakage of membership information for low values of $n$, while this drops almost to the value of a random guess for large values ($n=8000$).

%%% Local Variables:
%%% mode: latex
%%% TeX-master: "main"
%%% End:

%% file: plots/Gauss.tex
\begin{tikzpicture}
  \coordinate (A) at (0cm,0cm);
  \coordinate (B) at (0cm,-1.1\standaloneHeight);
  \begin{axis}[at=(A), anchor=north west,
    axis2, width = \standaloneWidth,
    height = \standaloneHeight, ylabel={LB: \ref*{pgf:LB11}, SR: \ref*{pgf:SR11}, UB: \ref*{pgf:UB11}},
    ]
    \addplot +[] table[basicTable,x=b,y=a]{./plots/Gaussian/SucRateVsGenD20.csv};
    \label{pgf:SR11}
    \addplot +[] table[basicTable,x=b,y=a]{./plots/Gaussian/LBVsGenD20.csv};
    \label{pgf:LB11}
    \addplot +[] table[basicTable,x=b,y=a]{./plots/Gaussian/UBVsGenD20.csv};
    \label{pgf:UB11}
  \end{axis}

  \begin{axis}[at=(B), anchor=north west,
    leftAxis1,  width = \standaloneWidth,
    height = \standaloneHeight, ylabel={MI: \ref*{pgf:MI12}},
    ]
    \addplot+[] table[basicTable, y = a, x = b ]{./plots/Gaussian/MIsGenD20.csv};
    \label{pgf:MI12}
  \end{axis}
  \begin{axis}[at=(B), anchor=north west,
    rightAxis1,  width = \standaloneWidth,
    height = \standaloneHeight, ylabel={UB: \ref*{pgf:UB12}}]
    \pgfplotsset{cycle list shift=+3}
    \addplot+[] table[basicTable, y = a, x = b ]{./plots/Gaussian/UBVsGenD20.csv};
    \label{pgf:UB12}
  \end{axis}
\end{tikzpicture}

%% file: plots/attacks.tex
\begin{tikzpicture}
  \coordinate (A) at (0cm,0cm);
  \coordinate (B) at (0cm,-1.1\standaloneHeight);
  \coordinate (C) at (0cm,-2.2\standaloneHeight);

  % CIFAR10
  \begin{semilogxaxis}[at=(A), anchor=north west,
    leftAxis,
    width = \standaloneWidth,
    height = \standaloneHeight,
    ylabel = {},
    ]
    \addplot+[error bars/.cd, y dir=both, y explicit] table[basicTable, y = a, x = b, y error= c ]{./plots/Cifar10/PsucLikihood_1.csv};
    \label{pgf:Like}
    \addplot+[error bars/.cd, y dir=both, y explicit] table[basicTable, y = e, x = b, y error= f ]{./plots/Cifar10/PsucLikihood_1.csv};
    \label{pgf:Loss}
    \addplot+[error bars/.cd, y dir=both, y explicit] table[basicTable, y = g, x = b, y error= h ]{./plots/Cifar10/PsucLikihood_1.csv};
    \label{pgf:Ment}
    % \addlegendentry{Likelihood Attack}
    \addplot+[error bars/.cd, y dir=both, y explicit] table[basicTable, y = a, x = b, y error= c ]{./plots/Cifar10/PSucLB2_1.csv};
    \label{pgf:LB}
    % \addlegendentry{Lower Bound}
  \end{semilogxaxis}
  \begin{semilogxaxis}[at=(A), anchor=north west,
    rightAxis,
    width = \standaloneWidth,
    height = \standaloneHeight,
    ytick={0.2,0.4,0.6},
    yticklabels={.2,.4,.6}, ymin = 0.0, ymax = .8, 
    ylabel = {},
    ]
    \pgfplotsset{cycle list shift=+3}
    \addplot+[error bars/.cd, y dir=both, y explicit] table[basicTable, y = a, x = b, y error= c ]{./plots/Cifar10/Acc_1.csv};
    \label{pgf:Acc}
  \end{semilogxaxis}

  % MNIST
  \begin{semilogxaxis}[at=(B), anchor=north west,
    leftAxis,
    width = \standaloneWidth,
    height = \standaloneHeight,
    ylabel={LB: \ref*{pgf:LB1}, LK: \ref*{pgf:Like1}, LS: \ref*{pgf:Loss1}, ME: \ref*{pgf:Ment1}}, ymin=0.4, ymax=.8, ytick={0.5,0.6,0.7}, yticklabels={0.5,0.6,0.7},
    ]
    \addplot+[error bars/.cd, y dir=both, y explicit] table[basicTable, y = a, x = b, y error= c ]{./plots/MNIST/PsucLikihood_1.csv};
    \label{pgf:Like1}
    \addplot+[error bars/.cd, y dir=both, y explicit] table[basicTable, y = e, x = b, y error= f ]{./plots/MNIST/PsucLikihood_1.csv};
    \label{pgf:Loss1}
    \addplot+[error bars/.cd, y dir=both, y explicit] table[basicTable, y = g, x = b, y error= h ]{./plots/MNIST/PsucLikihood_1.csv};
    \label{pgf:Ment1}
    % \addlegendentry{Likelihood Attack}
    \addplot+[error bars/.cd, y dir=both, y explicit] table[basicTable, y = a, x = b, y error= c ]{./plots/MNIST/PsucLB2_1.csv};
    \label{pgf:LB1}
    % \addlegendentry{Lower Bound}
  \end{semilogxaxis}
  \begin{semilogxaxis}[at=(B), anchor=north west,
    rightAxis,
    width = \standaloneWidth,
    height = \standaloneHeight,
    ylabel={Acc: \ref*{pgf:Acc1}}, ymin=0.0, ymax=1, ytick={0.25,0.5,0.75}, yticklabels={0.25,0.5,0.75},
    ]
    \pgfplotsset{cycle list shift=+3}
    \addplot+[error bars/.cd, y dir=both, y explicit] table[basicTable, y = a, x = b, y error= c ]{./plots/MNIST/Acc_1.csv};
    \label{pgf:Acc1}
  \end{semilogxaxis}

  % Fashion MNIST
  \begin{semilogxaxis}[at=(C), anchor=north west,
    leftAxis,
    width = \standaloneWidth,
    height = \standaloneHeight,
    ylabel = {},
    ]
    \addplot+[error bars/.cd, y dir=both, y explicit] table[basicTable, y = a, x = b, y error= c ]{./plots/FashionMNIST/PsucLikihood_1.csv};
    \label{pgf:Like2}
    \addplot+[error bars/.cd, y dir=both, y explicit] table[basicTable, y = e, x = b, y error= f ]{./plots/FashionMNIST/PsucLikihood_1.csv};
    \label{pgf:Loss2}
    \addplot+[error bars/.cd, y dir=both, y explicit] table[basicTable, y = g, x = b, y error= h ]{./plots/FashionMNIST/PsucLikihood_1.csv};
    \label{pgf:Ment2}
    % \addlegendentry{Likelihood Attack}
    \addplot+[error bars/.cd, y dir=both, y explicit] table[basicTable, y = a, x = b, y error= c ]{./plots/FashionMNIST/PsucLB2_1.csv};
    \label{pgf:LB2}
    % \addlegendentry{Lower Bound}
  \end{semilogxaxis}
  \begin{semilogxaxis}[at=(C), anchor=north west,
    rightAxis,
    width = \standaloneWidth,
    height = \standaloneHeight,
    ylabel = {},
    ]
    \pgfplotsset{cycle list shift=+3}
    \addplot+[error bars/.cd, y dir=both, y explicit] table[basicTable, y = a, x = b, y error= c ]{./plots/FashionMNIST/Acc_1.csv};
    \label{pgf:Acc2}
  \end{semilogxaxis}
\end{tikzpicture}

%% file: plots/AttInf.tex
\begin{tikzpicture}%
  \coordinate (A) at (0cm,0cm);
  \coordinate (B) at (0cm,-1.1\standaloneHeight);
  
  \begin{semilogxaxis}[at=(A), anchor=north west,
    AttAxis, ylabel={Success Rate},
    ybar=0pt, bar width=1.04,
    attInfDef,
    ] % ylabel={A: \ref*{pgf:InfA}, B: \ref*{pgf:InfB}, C = \ref*{pgf:InfC}, D = \ref*{pgf:InfD}}]
    \addplot+[error bars/.cd, y dir=both, y explicit] table[basicTable, y = c, x = a, y error= d ]{./plots/PenDigits/penDigAttInfA.csv};
    \label{pgf:InfA}
    \addplot+[error bars/.cd, y dir=both, y explicit] table[basicTable, y = c, x = a, y error= d ]{./plots/PenDigits/penDigAttInfB.csv};
    \label{pgf:InfB}
    \addplot+[error bars/.cd, y dir=both, y explicit] table[basicTable, y = c, x = a, y error= d ]{./plots/PenDigits/penDigAttInfC.csv};
    \label{pgf:InfC}
    \addplot+[error bars/.cd, y dir=both, y explicit] table[basicTable, y = c, x = a, y error= d ]{./plots/PenDigits/penDigAttInfD.csv};
    \label{pgf:InfD}
    \addlegendentry{Confidence}
    \addlegendentry{Accuracy}
    \addlegendentry{Loss}
    \addlegendentry{Gradient}
  \end{semilogxaxis}

  \begin{semilogxaxis}[at=(B), anchor=north west,
    leftAxis3, ylabel={LB: \ref*{pgf:LB5}, SR: \ref*{pgf:Like5}},
    attInfDef,]
    \addplot+[error bars/.cd, y dir=both, y explicit] table[basicTable, y = c, x = a, y error= d ]{./plots/PenDigits/penDigMemInfLike.csv};
    \label{pgf:Like5}
    % \addlegendentry{Likelihood Attack}
    \addplot+[error bars/.cd, y dir=both, y explicit] table[basicTable, y = c, x = a, y error= d ]{./plots/PenDigits/penDigMemInfLB.csv};
    \label{pgf:LB5}
    % \addlegendentry{Lower Bound}
  \end{semilogxaxis}
  \begin{semilogxaxis}[at=(B), anchor=north west,
    rightAxis3, ylabel={Acc: \ref*{pgf:Acc5}},
    attInfDef,]
    \pgfplotsset{cycle list shift=+3}
    \addplot+[error bars/.cd, y dir=both, y explicit] table[basicTable, y = c, x = a, y error= d ]{./plots/PenDigits/penDigAcc.csv};
    % \addplot+[no marks] table[basicTable, y = c, x = a ]{./plots/PenDigits/penDigAcc.csv};
    \label{pgf:Acc5}
  \end{semilogxaxis}%
\end{tikzpicture}%

%% file: conclusion.tex
We proposed a theoretical framework to analyze and bound information leakage from machine learning models. Our framework allows us to draw strong privacy by drawing upper bounds on the success rate of \glspl{mia}. Furthermore, we studied how much information is stored in a trained model about its training data and the implications in terms of leakage of sensitive attributes from the model parameters.

Our numerical experiments illustrate how the bounds from \cref{sec:GenErr} can be used to assess the information leakage of \gls{ml} models. The success rate of the attacker follows the same behavior as its lower bound, which is a function of the generalization gap. As a lower bound, it cannot guarantee that there is no attack that can perform better under the same conditions. Nevertheless, if the lower bound is above the performance of a random guess, the target model is guaranteed to leak sensitive information about its training data; thus, the generalization gap can be used to alert that a model is leaking sensitive information.

We exposed the leakage of sensitive information via attribute inference attacks, proving that models that are susceptible to membership inference can also be susceptible to other, more severe, privacy violations. We collected several attribute inference strategies and compared their effectiveness, finding that there is gain to be had by fully exploiting white-box access to the model.

The success rate of the Bayesian attacker provides a strong guarantee on the privacy of a model. However, computing the associated decision region seems computationally infeasible. In this paper we provided a synthetic example, using linear regression and Gaussian data, in which it is possible to analytically compute the involved distributions. In future work, we will explore novel tools to extend our illustrative examples to a systematic analysis of complex models. We hope that the present work serves as a common framework to compare different inference attack strategies. Furthermore, we hope that the definition of the Bayesian attacker and its connection to the generalization gap and the information stored by the model serve as inspiration to devise novel attack strategies and defense mechanisms, when applied to specific models.

%% file: appendix.tex
\section{Experimental Details}
\label{app:details}

Most of the experiments were run on a Latitude-7400 computer, with an Intel Core i7-8665U CPU @ 1.90GHz x 8 Processor. Part of the experiments were run on a server with two NVIDIA Quadro RTX 6000 GPUs and an AMD EPYC 7302 16-Core processor. 

The code and instructions necessary to reproduce our experiments can be found at \url{https://github.com/anonymus369/Formalizing-Attribute-and-Membership-Inference}.

\subsection{Examples on DNNs}
\label{app:details:DNNs}

The number of samples in the training set, $n$, varies in our experiments. For fixed $n$, that many samples are uniformly randomly picked from a pool of training samples. A test set is also fixed to measure the accuracy of the trained model and to empirically compute the generalization gap. In the case of MNIST and MNIST fashion, the training set is picked from a pool of $60$k samples. A separate pool of $10$k samples is fixed as the test set. For CIFAR10, the pool of training samples is of size $50$k, and the pool of test samples is of size $10$k.

The target model in this setup is a Deep Neural Network with $4$ convolutional layers and $3$ fully connected layers. For CIFAR10 the model has a total of $439722$ parameters, while for MNIST and MNIST fashion it has only $376714$. The model is trained for up to $150$ epochs using the Adam optimizer~\cite{kingma2014method} with learning rate $5\cdot 10^{-3}$. The batch size used for training the models is $200$ (this represent the whole training set when $n\leq200$). An early stop criteria compares the current loss over the training set to the total loss after the previous epoch, and stops training if the difference in below $10^{-3}$. The number of epochs of training can change drastically depending on the size of the training set.

Regarding the likelihood attack, note that \Cref{alg:attak1} outputs $1$ if the attacker infers membership correctly and $0$ otherwise. The success rate of the attacker is computed by simply counting the number of times it succeeds, over the total number of trials. Experimentally, we found that a threshold of $h=0.8$ works best across different values of $n$.

For each $n$, we repeat the following process $100$ times: Draw uniformly the training set, train the model, compute the generalization gap, compute the lower bound on the success rate of the Bayesian attacker and perform the likelihood attack $10000$ times. The results over different realizations of the model are averaged to produce a single value for each $n$. 

\subsection{Attribute Inference on PenDigits}
\label{app:details:PenDig}

The PenDigits dataset~\cite{Dua:2019}, was taken by asking participants to write digits from $0$ to $9$ on a tablet. The original data contains variable-length time series that correspond to the position of the pen on the tablet over time. We pre-process the data to make the length of the time series uniform (length 32). Since the capture rate of the tablet is fixed, we can infer the time that it took to write a digit by the length of the original series. We keep this information, along with the number of strokes that were used to write the digit.

The target model is a \gls{dnn} trained to classify hand-written digits. The input to the network consist of two time series (one for each coordinate) indicating the position of the pen over time, an integer indicating the number of strokes, a float between $0$ and $1$ indicating the length of the original sequences and a one-hot-encoding of the identity of the writer. The latter is considered as the sensitive attribute, while the other inputs are considered non-sensitive.

The target model possesses $4$ fully-connected layers and a total of $4650$ parameters. The loss for training is the MSE between the soft probabilities and the one-hot-encoded labels; this is a bounded loss function, allowing us to use \cref{thm:bounded_loss} to lower bound the success rate of the Bayesian attacker. The model is trained with Adam optimizer (learning rate $5\cdot 10^{-3}$) for up to $2500$ epochs. An early stop criteria compares the current loss over the training set to the total loss after the previous epoch, and stops training if the difference is below $10^{-4}$.

In our experiments, we compute the success rate for each of the proposed attribute inference strategies as a function of the number of samples in the training set of the target model. For each value of $n$, we randomly uniformly select $100$ different training sets drawn from a pool containing a total of $11990$ samples. For each training set we train a model. Subsequently, we apply each attack criteria to $100$ training samples of each trained model. The success rate of the attacker is computed by counting the amount of times the attack is successful. The reported success rate is an average over different target models. Since there are $44$ different writers in the data set, a random guess would amount to a success rate of approximately $2.3\%$. 

Regarding membership inference, we compute the success rate of the likelihood attacker (\cref{alg:attak1}), the lower bound from \cref{thm:bounded_loss}, and the accuracy of the model on the test set as a function of $n$. For each value of $n$ we train $100$ different models and perform the attack $100$ times. Results are averaged over different models for each value of $n$. 

\section{Proof of Proposition~\ref{prop:1}}\label{Proposition-NP}
\label{app:prop1}

We  recall the definition of the total variation distance when applied to distributions $P$, $Q$ on a set $\mathcal{X}\subseteq \mathbb{R}^d $ and Scheffé's identity \cite[Lemma~2.1]{10.5555/1522486}
\begin{equation}
  \label{eq:scheffe}
\| P - Q\|_\text{TV} \triangleq \sup_{\mathcal{A}\in  \mathcal{B}^d} |P(\mathcal{A}) - Q(\mathcal{A})|=\frac12 \int |p_{X}(\mathbf{x})- q_{X}(\mathbf{x})|d\mu(\mathbf{x}) \,,
\end{equation}
with respect to a base measure $\mu$, where $\mathcal{B}^d$ denotes the class of all Borel sets on $\mathbb{R}^d$.

\begin{proof}
    First of all, we prove equality for $\gamma=1$. Let us denote the optimal decision regions with $\mathcal{T}^\star\equiv \mathcal{T}(1)$ and ${\mathcal{T}^\star}^c \equiv \mathcal{T}^c(1)$ (cf.\ \cref{def-decision-region}). Let   $\epsilon_{\text{0}}({\mathcal{T}^\star}^c)$ and $\epsilon_{\text{1}}(\mathcal{T}^\star )$ the Type-I and Type-II errors. Then, 
    \begin{align}
        & \epsilon_{\text{1}}(\mathcal{T}^\star  ) + \epsilon_{\text{0}}( {\mathcal{T}^\star}^c ) =  \int_{\mathcal{T}^\star}p_{\widehat{\theta}(\mathbf{Z})S|T}(\theta, s |0) \,d\theta ds + \int_{{\mathcal{T}^\star}^c}p_{\widehat{\theta}(\mathbf{Z})S|T}(\theta, s | 1) \,d\theta ds \nonumber\\
        &\qquad\qquad = \int_{\mathcal{T}^\star} \,\min_{t \in \{0,1\}} p_{\widehat{\theta}(\mathbf{Z})S|T}(\theta, s | t)\, d\theta ds \nonumber 
         + \int_{{\mathcal{T}^\star}^c} \,\min_{t \in \{0,1\}} p_{\widehat{\theta}(\mathbf{Z})S|T}(\theta, s | t)\,d\theta ds \nonumber \\
        &\qquad\qquad= \int_{\Theta \times \mathcal{S}} \,\min_{t \in \{0,1\}} p_{\widehat{\theta}(\mathbf{Z})S|T}(\theta, s | t)\, d\theta ds\nonumber\\
        &\qquad\qquad=	1- \left\| p_{\widehat{\theta}(\mathbf{Z})S|T}(\cdot | 1)-p_{\widehat{\theta}(\mathbf{Z})S|T}(\cdot| 0) \right\|_\text{TV} = 1 - \Delta ,\label{eq-last-id}
    \end{align}
    where the last identity follows by applying Scheffé's identity~\cref{eq:scheffe}. From \cref{eq-last-id}, we have for any decision region  $\widehat\TTT\subseteq \Theta \times \mathcal{S}$,
    \begin{align}
        1 - \Delta
        &= \int_{\Theta \times \mathcal{S}} \,\min_{t \in \{0,1\}} p_{\widehat{\theta}(\mathbf{Z})S|T}(\theta, s |  t) \,d\theta ds  \nonumber\\
        &= \int_{\widehat\TTT} \,\min_{t \in \{0,1\}} p_{\widehat{\theta}(\mathbf{Z})S|T}(\theta, s |  t) \,d\theta ds 
         + \int_{\widehat\TTT^c} \,\min_{t \in \{0,1\}} p_{\widehat{\theta}(\mathbf{Z})S|T}(\theta, s |  t) \,d\theta ds 
        \nonumber\\
        &\leq  \int_{\widehat\TTT} p_{\widehat{\theta}(\mathbf{Z})S|T}(\theta, s | 0) \,d\theta ds + \int_{\widehat\TTT^c}  p_{\widehat{\theta}(\mathbf{Z})S|T}(\theta, s | 1) \,d\theta ds 
        \nonumber \\
        & =  \epsilon_{\text{1}}(\widehat\TTT   ) + \epsilon_{\text{0}}( {\widehat\TTT}^c ).
    \end{align}
    
    It remains to show \cref{eq:prop1:2}, assuming that $\Prob\{T=1\}=\Prob\{T=0\}=1/2$. Using \cref{eq-last-id}, we have 
    \begin{align}
        \frac12\big[	1 - \Delta \big]   
        % & = \frac12\int_{\Theta \times \mathcal{S}} \,\min_{t \in \{0,1\}} \big\{p_{\widehat{\theta}(\mathbf{Z})S|T}(\theta, s |  t) \big\} d\theta ds 
        % \nonumber\\
        % & = \int_{\Theta \times \mathcal{S}} \,\min_{t \in \{0,1\}} \big\{p_{\widehat{\theta}(\mathbf{Z})S|T}(\theta, s |  t) \big\} d\theta ds
        % \nonumber\\
        % & = \mathbb{E}_{\widehat{\theta}(\mathbf{Z})S} \left[ \min_{t \in \{0,1\}}\Big \{p_{T|\widehat{\theta}(\mathbf{Z})S}( 0|\widehat{\theta}(\mathbf{Z}),S), p_{T|\widehat{\theta}(\mathbf{Z})S}(1|\widehat{\theta}(\mathbf{Z}),S) \Big\} \right]
        % \nonumber\\
        & =\frac12\big[ \epsilon_{\text{1}}(\mathcal{T}^\star ) + \epsilon_{\text{0}}( {\mathcal{T}^\star}^c )\big]\nonumber  \\
        & =  \inf_\varphi \Prob\left\{\varphi(\widehat{\theta}(\mathbf{Z}),S)\neq T \right\},
    \end{align}
    where the last identity follows by the definition of the decision regions. 
\end{proof}

\section{Proof of Theorem~\ref{thm:MSE_Exp}}
\label{appendix-ProofThm4}

\begin{proof}
  Using the definitions from the previous proofs, let $R \triangleq \varrho(\widehat\theta(\vt Z), \ol X, \ol Y)$ be the square of a sub-Gaussian random variable $\RSG\triangleq\sqrt{|R|}$ with variance proxy $\sigma_R^2$. Then, we have $\Prob\{R\ge r^2\}=\Prob\{|\RSG| \ge r \} \le 2 e^{-\frac{r^2}{2\sigma_R^2}}$ for all $r \ge 0$, which in turn yields $\Prob\{R\ge r\} \le 2 e^{-\frac{r}{2\sigma_R^2}}$ for all $r \ge 0$. Define the random variable $R_0$ to have the distribution function $Q_0(r) \triangleq P\{R_0 \le r \} \triangleq 1 - 2 e^{-\frac{r}{2\sigma_R^2}}$ on its support $[r_0, \infty)$, where $r_0 = 2 \sigma_R^2 \log 2$, i.e., the \gls{pdf} of $R_0$ is $p_{R_0}(r) = \frac{1}{\sigma_R^2} e^{-\frac{r}{2\sigma_R^2}}$.

  Let $Q$ be the distribution function of $R$. Then, using the construction in the proof of  in \cite[Theorem~1.104]{Klenke2013Probability}, we can write $R = Q^{-1} \circ Q_0(R_0)$, where $Q^{-1}$ is the left continuous inverse of $Q$, noting that $Q_0$ is continuous.  The tail bound on $R$ then implies $Q(r) = 1 - P\{|R| \ge r \} \ge Q_0(r)$, which immediately yields $Q^{-1} \circ Q_0(r) \le r$.

  Following similar steps as for \cref{thm:MSE_gaussian}, for $\Rmax \ge r_0$, we get,
  \begin{align}
    \label{eq:6}
    \int_{|r| \ge \Rmax} |r| p_R(r) dr
    &= \int_{Q^{-1} \circ Q_0(r) \ge \Rmax} Q^{-1} \circ Q_0(r) p_{R_0}(r)dr 
    \nonumber\\
    &= \int_{ Q_0(r) \ge Q(\Rmax)} Q^{-1} \circ Q_0(r) p_{R_0}(r)dr 
    \nonumber\\
    &\le \int_{ Q_0(r) \ge Q(\Rmax)} r p_{R_0}(r)dr 
    \nonumber\\
    &\le \int_{r \ge \Rmax} r p_{R_0}(r)dr 
    \nonumber\\
    &= \int_{\Rmax}^{\infty} \frac{r}{\sigma_R^2} e^{-\frac{r}{2\sigma_R^2}} dr 
    \nonumber\\
    &= -2\int_{\Rmax}^{\infty} \frac{\partial \;r e^{-\frac{r}{2\sigma_R^2}}}{\partial r}  dr + 2\int_{\Rmax}^{\infty}  e^{-\frac{r}{2\sigma_R^2}} dr
    \nonumber\\
    &= 2\Rmax\exp\left(-\frac{\Rmax}{2\sigma_R^2}\right) \left(1 + \frac{2\sigma_R^2}{\Rmax} \right).
  \end{align}
  
  The rest of the proof follows identically to that of \cref{thm:MSE_gaussian} and yields,
    \begin{align}
    \label{eq:7}
        &\PS(\varphi) \ge \max\Bigg\{P_{\mathrm{m}},
         P_{\mathrm{m}} \Bigg( \frac{|\EE[\EG(\Alg, \vt Z)]|}{2 \Rmax} \nonumber\\
         &- \frac{1}{1-P_\mathrm{m}}\exp\Bigg(-\frac{\Rmax}{2\sigma_R^2}\Bigg) \Bigg(1 + \frac{2\sigma_R^2}{\Rmax} \Bigg) - 1\Bigg) + 1\Bigg\}\;.
    \end{align}
\end{proof}

\section{Proof of Theorem~\ref{the4}}
\label{appendix-Theorem-info}

Before we proceed with the proof of \cref{the4}, we provide a series of definitions and preliminary results.
\subsection{Basic Definitions and Change of Measure}

Let us consider two probability measures $P$ and $Q$ on a common measurable space $(\Omega, \mathcal{F})$. Let $X$ denote a random variable $X:\Omega \rightarrow \mathcal{X}$ and $P_X$, $Q_X$ correspond to the induced distributions. Assuming absolute continuity $P_X\ll Q_X$, the KL-divergence of $Q_X$ with respect to $P_X$ is defined by
\begin{equation}
D_{\text{KL}}(P_X\|Q_X)\triangleq \mathbb{E}_{Q_X} \left [ -\log \left ( \frac{dP_X}{dQ_X}\right ) \right ].
\end{equation}
Consider a kernel  (or channel) according to  the law $P_{Y|X}$ that produces the random variable $Y$ given $X$. Let $P_Y$ be the induced distribution of $Y$ when $X$ is generated according to $P_X$ while  $Q_Y$ is the distribution of $Y$ when $X$ is generated according to $Q_X$. Then, by the data-processing inequality for KL-divergence~\cite[Theorem~2.2~6.]{Polyanskiy2014Lecture}, we have
\begin{equation}
D_{\text{KL}}(P_X\|Q_X) \geq D_{\text{KL}}(P_Y\|Q_Y).  \label{eq-data-processing}
\end{equation}
Equality holds if and only if $P_{X|Y}=Q_{X|Y}$, where $P_{X|Y} P_Y = P_{Y|X} P_X$ and $Q_{X|Y} Q_Y = P_{Y|X} P_X$. A simple application of this inequality leads to the following result. 

\begin{lemma}[Data-processing reduces KL-divergence~\cite{10.5555/1146355}]\label{lemma-data}
For any measurable set $\mathcal{B}\in  \mathcal{F}(\mathcal{X})$,  inequality \cref{eq-data-processing} applied to the degenerate channel based on the indicator function  $Y=\mathds{1}[X\in \mathcal{B}]$ implies: 
\begin{align}
D_{\text{KL}}(P_X\|Q_X) &\geq d_{\text{KL}}( p_\mathcal{B} \|q_\mathcal{B} ) = d_{\text{KL}}( 1-p_\mathcal{B} \|1-q_\mathcal{B} ),\label{eq.KL}
\end{align}
where $d_{\text{KL}}(\cdot \| \cdot  )$ denotes the binary KL-divergence  with parameters $p_\mathcal{B}=P_X(\mathcal{B})$ and $q_\mathcal{B}=Q_X(\mathcal{B})$. Note that if $t \in [0, 1]$ and $M > 1$, then
\begin{equation}
  \log_2(M) - d_{\text{KL}}(t \| 1- 1/M)= t \log_2(M-1) + H_2(t)\;,
\end{equation}
where $H_2(t) \triangleq -t\log_2 t - (1-t)\log_2(1-t)$ is the binary entropy function.
Equality holds in \cref{eq.KL} if and only if $P_{X|X\in \mathcal{B}} = Q_{X|X\in \mathcal{B}}$ and $P_{X|X\notin  \mathcal{B}} = Q_{X|X\notin  \mathcal{B}}$.
\end{lemma}
The proof of this lemma is rather straightforward from basic properties and will be omitted. We next revisit a well-known result to  obtain bounds for the probability of an arbitrary event $\mathcal{B} \in \mathcal{F}(\mathcal{X})$.

\begin{lemma}[Change of measure~\cite{bassily2018learners}]
\label{one}
Let the distributions $P_X$ and $Q_X$ be induced by the random variable $X$ as described. Then, 
\begin{equation}
\sup_{\mathcal{B} \in \mathcal{F}(\mathcal{X})} P_X(\mathcal{B})\log_2 (1/Q_X(\mathcal{B})) \leq D_{\text{KL}}(P_X\|Q_X)+1,  \label{eq-change-of-meausure1}
\end{equation}
where the supremum is taken over all measurable sets $\mathcal{B} \in \mathcal{F}(\mathcal{X})$. 
\end{lemma}
\begin{proof}
For any set $\mathcal{B} \in \mathcal{F}(\mathcal{X})$, we have by \cref{lemma-data}, that
\begin{align}
D_{\text{KL}}(P_X \| Q_X) &\geq d_{\text{KL}}( p_\mathcal{B} \|q_\mathcal{B} )
\nonumber\\
&= P_X(\mathcal{B})\log  \frac{P_X(\mathcal{B})}{Q_X(\mathcal{B})} +P_X(\mathcal{B}^c)\log  \frac{P_X(\mathcal{B}^c)}{Q_X(\mathcal{B}^c)} 
\nonumber\\
&=P_X(\mathcal{B})\log_2 \frac{1}{Q_X(\mathcal{B})} +P_X(\mathcal{B}^c)\log_2  \frac{1}{Q_X(\mathcal{B}^c)}-H_2(p_\mathcal{B}) 
\nonumber\\
&\geq P_X(\mathcal{B})\log_2 \left ( \frac{1}{Q_X(\mathcal{B})}\right )-1 . \label{eq-change-of-meausure2}
\end{align}

The final inequality \cref{eq-change-of-meausure1} follows by taking the supremum over all measurable sets $\mathcal{B} \in \mathcal{F}(\mathcal{X})$ in \cref{eq-change-of-meausure2}.  
\end{proof}

\subsection{Cram\'er-Chernoff Method}
\label{app:cramer-chernoff}
We recall a distribution-dependent deviation  bound based on the optimization of the Markov inequality which is known as Cram\'er-Chernoff method.

Let $Z$ be a real-valued random variable and define its log-moment-generating function as
\begin{equation}
 \psi_{Z}(\lambda)=\log  \mathbb{E}\left [ \exp{\lambda Z} \right ],\ \ \lambda\geq 0.
\end{equation}
 For $\lambda \ge 0$, the Markov inequality implies:
\begin{align}
  \Prob(Z \geq t) &\le \Prob(e^{\lambda Z} \ge e^{\lambda t}) \nonumber\\  % \le here is due to the possbility that lambda = 0
                  &\le e^{-\lambda t} \mathbb{E}[e^{\lambda Z}] \nonumber\\
                   &  =\exp\big[ -\lambda t + \psi_{Z}(\lambda) \big] . \label{eq-markov-bound1} 
\end{align}
As \cref{eq-markov-bound1} holds for any $\lambda \ge 0$, we immediately obtain $\Prob(Z \geq t) \leq \exp \left [-\psi_{Z}^{\ast}(t) \right ]$ for $t \ge \mathbb{E}[Z]$, where 
\begin{equation}
 \psi_{Z}^{\ast}(t)=\sup_{\lambda \in \RR}\limits\left \{\lambda t -\log  \mathbb{E}\left [ e^{\lambda Z} \right ] \right \}. \label{eq-dual}
\end{equation}
This expression is known as the Fenchel-Legendre dual function of $\psi_{Z}(\lambda)$ and it equals 
$\psi_{Z}^{\ast}(t) = \sup\limits\left \{\lambda t -\psi_{Z}(\lambda) \,:\, \lambda \geq 0 \right \}$ whenever $t \geq \mathbb{E}[Z]$.

% Assuming $\psi_{Z}(\lambda) <\infty$ then $\psi_{Z}(\lambda)$ is convex in an open interval $I \ni \lambda$ and if $Z$ is centered then we can differentiate to obtain the solution: 
% \begin{align}
% \psi_{Z}^{\ast}(t) & =\sup_{\lambda \in I} \limits\left \{\lambda t -\log  \mathbb{E}\left [ e^{\lambda Z} \right ] \right \} 
% \nonumber\\
%  & =t \cdot (\psi_{Z}')^{-1}(t) - \psi_{Z}[(\psi_{Z}')^{-1}(t)].
     %   \end{align}

And therefore, for $t \ge \mathbb{E}[Z]$,
\begin{equation}
  \label{eq:9}
  \Prob(Z \geq t) \le \exp \left [-\psi_{Z}^{\ast}(t) \right ] .
\end{equation}

We will need the following properties: 
\begin{itemize}
\item If $Z=X_1+\dots +X_n$ with $\{X_i\}_{i=1}^n$ being i.i.d.\ copies of $X$, then 
\begin{align}
\psi_{Z}^{\ast}(t) =n\psi_{X}^{\ast}\left(\frac{t}{n} \right );
\label{fenc1}
\end{align}
\item For any random variable $Z$,
 \begin{align}
\psi_{Z/n}^{\ast}(t) =\psi_{Z}^{\ast}\left(nt \right ).
\label{fenc2}
\end{align}
\end{itemize}
An immediate consequence of these properties is that the  random variable 
$$
Z=  \mathbb{E}[X] -\frac{1}{n}\sum_{i=1}^n X_i  ,
$$ 
with $\{X_i\}_{i=1}^n$ i.i.d.\ copies of $X$, satisfies 
\begin{align}
    \Prob( Z \geq t)
    &\le \exp \left [-n\psi_{\mathbb{E}[X] -X}^{\ast}(t ) \right ] 
    ,\ \forall t \geq 0.
\label{concentration-bound}
\end{align}

\subsection{Proof}
Using these preliminary results, we are now ready to prove \cref{the4}.
The proof requires three steps which are described below. 

\textbf{Information loss.} First, we observe that $(T,S) \leftrightarrow  \mathbf{Z} \leftrightarrow   \widehat{\theta}(\mathbf{Z})$ and thus $T \leftrightarrow  (\mathbf{Z},S) \leftrightarrow   \widehat{\theta}(\mathbf{Z})$ form Markov chains since $\widehat{\theta}$ is a stochastic function of $\mathbf{Z}$. As a consequence of the data-processing inequality~\cite[Theorem~2.8.1]{10.5555/1146355}, we obtain \cref{eq-inequality-missing3} from
\begin{equation}
  I(T; \widehat{\theta}(\mathbf{Z}) | S) \leq   I(\mathbf{Z}; \widehat{\theta}(\mathbf{Z}) | S) = I(\mathbf{Z}; \widehat{\theta}(\mathbf{Z}) )  - I( S;\widehat{\theta}(\mathbf{Z}) ) .   
\end{equation}
Interestingly, we will show that $ I(T; \widehat{\theta}(\mathbf{Z}) | S)$ bounds the accuracy of the membership (sensitive attribute) inference while $I(\mathbf{Z}; \widehat{\theta}(\mathbf{Z}) )$ bounds the generalization gap of the hypothesis associated with $\widehat{\theta}$.  

\textbf{Generalization gap.} 
The proof of the bound on the generalization gap~\cref{eq-inequality-missing2} easily follows from application of well-known results. For $\epsilon \ge 0$ let us define the region
  \begin{align}
    \mathcal{B} &\triangleq  \left\{ (\theta,\TrainSet) \in \Theta\times \mathcal{Z}^n : \EE[\varrho(\theta, Z)] - \frac{1}{n} \sum_{i=1}^n \varrho(\theta, z_i) \ge \epsilon  \right\}.
  \end{align}
  By the definition of the generalization gap (\cref{def:expected_risks}), we have $\GG(\Alg, \vt Z) \ge \epsilon$ if and only if $\big(\widehat{\theta}(\mathbf{Z}), \mathbf{Z}\big) \in \mathcal{B}$.
  We define the associated fibers $\mathcal{B}_{(\theta)} \triangleq \{\TrainSet \in \mathcal{Z}^n : (\theta,\TrainSet) \in \mathcal{B}\}$ for $\theta \in \Theta$.
  First, we apply the Cram\'er-Chernoff method~(\cref{app:cramer-chernoff}) to the random variable
  \begin{align}
    R_\theta \triangleq  \mathbb{E}[\varrho({\theta},(X,Y))] - \varrho({\theta},(X,Y))
  \end{align}
  $\mathcal{B}_{(\theta)}$ with respect to the data probability measure $P_{\mathbf{Z}}$, where \cref{eq:9,fenc1,fenc2} then yield
  \begin{align}
    \Prob\left(\mathbf{Z}\in \mathcal{B}_{(\theta)}\right) 
    &\le \exp \big[-n \psi_{R_\theta}^{\ast}(\epsilon)\big],
    \label{eq-fenc-bound2}
\end{align}
where $\psi_{R_\theta}^{\ast}$ is the Fenchel-Legendre dual function of the real-valued random variable: . Then, it follows that, 
\begin{align}
\esssup_{\theta \sim P_{\widehat\theta(\mathbf{Z})}} \Prob\left(\mathbf{Z}\in \mathcal{B}_{(\theta)}\right)&  \leq  \exp \Big[ -n\essinf_{\theta \sim P_{\widehat\theta(\mathbf{Z})}}  \psi_{R_\theta}^{\ast}(\epsilon)\Big ]. 
\label{eq-123-B}
\end{align}
We can now use \cref{one}, rearranging terms and taking the expectation w.r.t.\ $P_{\widehat\theta(\mathbf{Z})}$, we have that
\begin{align}
    \Prob\left( \GG(\Alg, \mathbf{Z}) \geq \epsilon  \right)& = \Prob\left((\widehat{\theta}(\mathbf{Z}), \mathbf{Z})\in \mathcal{B} \right)
                                                              \nonumber\\
                                                            & \leq \frac{I(\mathbf{Z}; \widehat{\theta}(\mathbf{Z}) )+1}{- \log(P_{\widehat\theta(\mathbf Z)} \times P_{\mathbf Z}(\mathcal B))} 
                                                              \label{eq:lemma2bound}\\
                                                            & \leq \frac{I(\mathbf{Z}; \widehat{\theta}(\mathbf{Z}) )+1}{- \log\big(\int P_{\mathbf Z}(\mathcal B_{(\theta)}) \, dP_{\widehat\theta(\mathbf Z)}(\theta) \big)} 
                                                              \nonumber\\
                                                            & \leq \frac{I(\mathbf{Z}; \widehat{\theta}(\mathbf{Z}) )+1}{-\log\Big[  {\esssup\limits_{\theta \sim P_{\widehat\theta(\mathbf{Z})}}} \, \mathbf{P}\big(\mathbf{Z}\in \mathcal{B}_{(\theta)}\big) \Big] } 
                                                              \nonumber\\
                                                            & \leq \frac{I(\mathbf{Z}; \widehat{\theta}(\mathbf{Z}) )+1}{n\Big[ \essinf\limits_{\theta \sim  P_{\widehat\theta(\mathbf{Z})}}  \psi_{R_\theta}^{\ast}(\epsilon)  \Big]\label{eq-123-A}
                                                              },
\end{align}
where inequality \cref{eq:lemma2bound} follows from \cref{eq-change-of-meausure1} and \cref{eq-123-A} follows from \cref{eq-123-B}. 

\textbf{Attribute inference.}  Let $\varphi^\star$ be the attack strategy given in~\cref{eq:thm1:2} with $\PS(\varphi^\star )= \Prob\{\widehat T = T\} \ge 1/2$, where $\widehat T$ denotes the random variable $\widehat T \triangleq \varphi^\star (\widehat\theta(\vt Z), S)$. Note that $\widehat T$ is independent of $T$ given $(\widehat\theta(\vt Z), S)$. We will show that,
\begin{align}
  I(T; \widehat{\theta}(\mathbf{Z}) | S) &\geq 
    d_{\text{KL}}\left ( \PS(\varphi^\star) \, \middle\| \,  \mathbb{E}\Big[\min_{t\in\mathcal{T}} P_{T|S} (t|S)  \Big] \right),
\end{align}
where,
\begin{align}
d_{\text{KL}}( p \|q ) &\triangleq  p \log_2  \frac{p }{q} + (1-p)\log_2  \frac{(1-p)}{(1-q)}. 
\end{align}
To this end, denote by $D_{\text{KL}}(\cdot \| \cdot)$ the KL-divergence between two distributions and observe that, by \cref{lemma-data},
\begin{equation}
    D_{\text{KL}} \left( P_{T|\widehat{\theta}(\mathbf{Z}) S}(\cdot|\theta, s) \| P_{T|S} (\cdot|s) \right)\geq  d_{\text{KL}}\left (  P_{T|\widehat{\theta}(\mathbf{Z}) S}\big(t | \theta, s\big) \, \big \| \,  P_{T|S} \big(t |s\big) \right) ,              
\end{equation}
\added{where $t = t(s, \theta)$ may be any function of $(s, \theta) \in \mathcal{S} \times \Theta$.}
By taking the expectation over $\theta, s \sim p_{S,{\widehat{\theta}(\mathbf{Z})}}$, we obtain
\begin{align}
  I(T; &\widehat{\theta}(\mathbf{Z}) | S) \geq \mathbb{E} \left[ d_{\text{KL}}\left (  P_{T|\widehat{\theta}(\mathbf{Z}) S}\big(t| \widehat{\theta}(\mathbf{Z}), S\big) \, \big \| \,  P_{T|S} \big(t|S\big) \right)\right] .
  \label{eq-mut_inf_bound}
\end{align}
We \added{choose} a mapping $t^\star_{(s,{\theta})}$ that satisfies,
\begin{equation}
\mathbb{E}\Big[ P_{T|\widehat{\theta}(\mathbf{Z}) S} (t^\star|\widehat{\theta}(\mathbf{Z}),S)  \Big] = \mathbb{E}\Big[\max_{t\in\mathcal{T}} P_{T|\widehat{\theta}(\mathbf{Z}) S} (t|\widehat{\theta}(\mathbf{Z}),S)  \Big] = \PS(\varphi^\star).
\end{equation}
It is straightforward to verify that,
\begin{align}
\PS(\varphi^\star) \geq \mathbb{E}\Big[\max_{t\in\mathcal{T}} P_{T|S} (t|S)  \Big].\label{eq-verify}
\end{align}
Then, by convexity of the function $(p,q)\mapsto d_{\text{KL}}( p \|q )$, we can continue from \cref{eq-mut_inf_bound} to show,
\begin{align}
  I(T; \widehat{\theta}(\mathbf{Z}) | S) &\geq  \mathbb{E} \left[ d_{\text{KL}}\Big ( P_{T|\widehat{\theta}(\mathbf{Z}) S}(t^\star | \widehat{\theta}(\mathbf{Z}), S) \big \| \,  P_{T|S} (t^\star |S) \Big) \right]
  \nonumber\\
  &\geq d_{\text{KL}}\left( \PS(\varphi^\star) 
    \, \big \| \, \mathbb{E}\big[ P_{T|S} \big(t^\star|S\big)\big] \right)\ 
    \nonumber\\
  &\geq d_{\text{KL}}\left( \PS(\varphi^\star) 
    \, \big \| \, \mathbb{E}\big[\max_{t\in\mathcal{T}} P_{T|S} (t|S)  \big] \right), \label{eq-mut_inf_bound2} 
\end{align}
where the last inequality~\cref{eq-mut_inf_bound2} follows by using~\cref{eq-verify} and noticing that the function $q\mapsto d_{\text{KL}}( p \|q )$ is non-increasing for $q\in[0,p]$.

Finally, notice that we can apply the bound,
 \begin{align}
d_{\text{KL}}( p \|q ) &\geq  \max\big\{ 2 (p - q)^2,  -p \log_2 (q) - 1\big\}\;,
\end{align}
with $p\geq  q$. 
\qed

\section{Gaussian Data and Linear Regression}
\label{appendix-GaussianExample}

Recall the following notation: $\vt x$ is the $[d \times n]$ matrix given by $\vt x = (x_1, x_2, \dots, x_n)$, while $\vt y = (y_1, y_2, \dots, y_n)$ and $\vt \Noise = (\Noise_1, \Noise_2, \dots, \Noise_n)$ are $[1 \times n]$ vectors. Let each copy of noise $\Noise$ be normal i.i.d.; $\Noise\sim\NNN({}\cdot{};0,\sigma^2)$. Since $Y_i$ is linear in $\Noise_i$, $Y_i$ is also normal distributed, $Y_i\sim\NNN({}\cdot{};\beta^T x_i, \sigma^2)$. Since model parameters are linear in the training set $\vt Y$, their pdf\ is a multivariate Gaussian, $\widehat\theta(\vt Y) \sim Q({}\cdot{})\triangleq\NNN({}\cdot{};\beta,\sigma^2 \ol x^{-1})$, where $\ol x \triangleq \vt x \vt x^T$. Furthermore, fixing the $j-\mathrm{th}$ sample in the training set to $s$, we have $\widehat\theta(\vt Y)$ distributed as $Q_{j}({}\cdot{}|s)\triangleq\NNN\big({}\cdot{}; \beta + \ol x^{-1}x_j(s - x_j^T \beta),\sigma^2\ol x^{-1}(\mathbb{I}^{d\times d} - x_j x_j^T \ol x^{-1})\big)$.

Consider a \gls{mia} against this model. The attacker possesses side information $(S_J,J)$, that is, a test sample and its corresponding index. Recall our definition $S_\RDI=T(Y_\RDI)+(1-T)(Y'_\RDI)$, where $\RDI$ is a random index in $[n]$. When $T=0$, $S=Y'_\RDI$, independent of the training set; hence,
\begin{align}
    p_{S_\RDI\RDI\widehat\theta|T}(s,j,\theta|0) &= \frac{1}{n} p_{S_\RDI \widehat\theta(\vt Y)|T,\RDI}(s,\theta|0,j) 
    \nonumber\\ &= \frac{1}{n}p_{\widehat\theta(\vt Y)|T}(\theta|0) p_{S_\RDI|T,\RDI}(s|0,j)  
    \nonumber\\
    &= \frac{1}{n} Q(\theta) p_{Y_j}(s) \;,
    \label{eq:4.1.1}
\end{align}

On the other hand, when $T=1$, $S = Y_\RDI$ is the $J-\mathrm{th}$ component of the training set $\vt Y$; therefore,
\begin{align}
     p_{S_\RDI\RDI\widehat\theta|T}(s,j,\theta|1) &= \frac{1}{n} p_{S_\RDI \widehat\theta(\vt Y)|T,\RDI}(s,\theta|1,j) 
     \nonumber\\
    &= \frac{1}{n} p_{\widehat\theta(\vt Y)|T\RDI S_\RDI}(\theta|1,j,s) p_{S_\RDI|T,\RDI}(s|1,j)
    \nonumber\\
    &= \frac{1}{n} Q_{j}(\theta|s) p_{Y_j}(s) \;.
    \label{eq:4.1.2}
\end{align}

Note that $Q({}\cdot{})$ and $Q_i({}\cdot{}|s)$ differ only by their mean and variance. The second pdf\ has shifted mean and reduced variance. The reduced variance is to be expected, since fixing one of the samples in the training set should reduce randomness. Note that if the dimension of the space of features is equal to the amount of samples (i.e., $d\geq n$) an attacker having access to the feature vectors in the training set $\vt x$ can solve a system of equations to obtain $\vt y$.

In the following, we derive a theoretical lower bound for \cref{eq:SuccProbGaussian1}. Define $R \triangleq x_\RDI^T(\vt x \vt x^T)^{-1}\vt x \vt Y^T - S_\RDI$. Fixing $\RDI$ and $T$, $R$ is a linear combination of Gaussian r.v.s, and thus $R$ is a Gaussian random variable. Regardless of $T$ and $\RDI$, $\EE[R] = 0$. If $T=0$; then $S_\RDI = Y'_\RDI$, independent of $\vt Y$,
\begin{align}
    \Var[R|T=0] = \sigma^2 + \frac{d\sigma^2}{n}\;.
\end{align}

If $T=1$, then $S_\RDI = Y_\RDI$ is the $J-\mathrm{th}$ component of $\vt Y$; consequently,
\begin{align}
    \Var[R|T=1] = \sigma^2 - \frac{d\sigma^2}{n}\;.
\end{align}

In total,
\begin{align}
    \sigma_R^2 \triangleq \Var[R] &= \sigma^2 \;.
\end{align}

Since $R$ is a Gaussian random variable, the squared error, defined by $R^2\triangleq\left(x^T(\vt x \vt x^T)^{-1}\vt x \vt Y - S_\RDI\right)^2$, is exponentially tail-bounded\footnote{See proof of \cref{thm:MSE_Exp} in \cref{appendix-ProofThm4}.}; hence, we can apply \cref{thm:MSE_Exp} to get a theoretical lower bound on the success probability of the Bayesian \gls{mia}. Assume that $T$ is Bernoulli $1/2$ distributed; thus,

\begin{align}
    \PS(\varphi^{\star}) &\ge \frac{1}{2} + \frac{|\EE[\EG(\Alg, \vt Z)]|}{4 \Rmax} 
    \nonumber\\
    &- \exp\left(-\frac{\Rmax}{2\sigma_R^2}\right) \left(1 + \frac{2\sigma_R^2}{\Rmax} \right) \;.
    \nonumber\\
    &= \frac{1}{2} + \frac{d}{2n }\frac{\sigma^2}{\Rmax} - \exp\left(-\frac{\Rmax}{2\sigma^2}\right) \left(1 + \frac{2\sigma^2}{\Rmax} \right)\;,
    \label{eq:SuccProbGausLB}
\end{align}
where we use \cref{eq:genErrorGaus1}.

The Mutual information between a test sample $S_\RDI$ and the model parameters $\widehat{\theta} (\vt Y)$ given the sensitive attribute $T$ is,

\begin{align}
    I(S_\RDI;\widehat{\theta} (\vt Y)|T) &=  \sum_{t\in\{0,1\}}I(S_\RDI;\widehat{ \theta}(\vt Y)|T=t)
    \nonumber\\
    &= \Prob\{T=1\}I(S_\RDI;\widehat{\theta}(\vt Y)|T=1)
    \nonumber\\
    &= \Prob\{T=1\}\frac{1}{n}\sum_{j=1}^n \int Q_j(\theta|s)P_{Y_j}(s)\log\left[\frac{Q_j(\theta|s)P_{Y_j}(s)}{Q(\theta) P_{Y_j}(s)}\right]d\theta ds
    \nonumber\\
    &=\Prob\{T=1\}\frac{1}{n}\sum_{j=1}^n \int Q_j(\theta|s)P_{Y_j}(s)\log\left[\frac{Q_j(\theta|s)}{Q(\theta) }\right]d\theta ds
    \nonumber\\
    &=\Prob\{T=1\}\frac{1}{n}\sum_{j=1}^n \EE_{S\sim P_{Y_j}}\left[D_{\mathrm{KL}(Q_j(\theta|s)|Q(\theta))}\right]
    \nonumber\\
    &= \Prob\{T=1\}\frac{1}{2n}\sum_{j=1}^n \EE_{S\sim P_{Y_j}}\Big[\Tr{\left(\Sigma^{-1}\Sigma_j\right)}-\log\left(\frac{|\Sigma_j|}{|\Sigma|}\right)-d
    \nonumber \\
    &\qquad +(\mu_j(S)-\beta)^T\Sigma^{-1}(\mu_j(S)-\beta)\Big]
    \nonumber\\
    &= \Prob\{T=1\}\frac{1}{2n}\sum_{j=1}^n \left(\Tr{\left(\Sigma^{-1}\Sigma_j\right)}-\log\left(\frac{|\Sigma_j|}{|\Sigma|}\right)-d+x_j^T\ol x^{-1} x_j\right)
    \nonumber\\
    &= \Prob\{T=1\}\frac{1}{2n}\sum_{j=1}^n\log\left(\frac{|\Sigma|}{|\Sigma_j|}\right)
    \;,
    \label{eq:MI2}
\end{align}

with $\mu_j(s)\triangleq \beta + \ol x^{-1}x_j(s - x_j^T \beta)$, $\Sigma_j\triangleq\sigma^2\ol x^{-1}(\mathbb{I}^{d\times d} - x_j x_j^T \ol x^{-1})\big)$ and, $\Sigma = \ol x^{-1}\sigma^2$. Using the upper bound $I(S_\RDI;\widehat{\theta} (\vt Y)|T)\geq I(T;\widehat{\theta} (\vt Y)|S_\RDI)$ in combination with \cref{eq-inequality-missing1}, we can estimate an upper bound on the probability of success of the Bayesian attacker.

\begin{proof}[Proof of~\cref{eq:genErrorGaus1}]
    Recall the definition of the generalization gap, substituting the MSE and the model into the definition, we obtain,
    \begin{align}
        \EE \left [    \EG(\Alg, \vt Y) \right]        &=\EE \left[\frac{1}{n}\sum_{i=1}^n\ell(f_{\widehat\theta(\vt Y)}(x_i),Y'_i)-\frac{1}{n}\sum_{i=1}^n\ell(f_{\widehat\theta(\vt Y)}(x_i),Y_i)\right]
        \nonumber\\
        &=\frac{1}{n}\EE \left[\norm{\widehat\theta(\vt Y)^T\vt x -\vt Y'}^2 - \norm{\widehat\theta(\vt Y)^T\vt x -\vt Y}^2\right]\;,
    \end{align}
    Let $\ol x\triangleq \vt x \vt x^T$, then,
    \begin{align}
        \EE \left[\norm{\widehat\theta(\vt Y)^T\vt x -\vt Y'}^2\right] &=\EE \left[\left(\vt Y \vt x^T \ol x ^{-1} \ol x \ol x^{-1} \vt x \vt Y ^T - 2 \vt Y' \vt x^T\ol x^{-1} \vt x \vt Y ^T + \norm{\vt Y'}^2\right)\right]
        \nonumber\\
        &=\EE \left[\vt \Noise \vt x^T \ol x^{-1} \vt x \vt \Noise ^T -\beta^T \ol x\beta - 2 \vt\Noise' x^T\ol x^{-1} \vt x \vt \Noise^T + \norm{\vt Y'}^2 \right]
        \nonumber\\
        &=\EE \left[\vt \Noise \vt x^T \ol x^{-1} \vt x \vt \Noise ^T -\beta^T \ol x\beta + \norm{\vt Y'}^2 \right]\;,
    \end{align}
    Note that $\EE \left[2 \vt\Noise' x^T\ol x^{-1} \vt x \vt \Noise^T\right]= 0$; since $\EE \left[\Noise\right]=0$ and $\vt\Noise'$ is independent from $\vt \Noise$. On the other hand,
    \begin{align}
        \EE \left[\norm{\widehat\theta(\vt Y)^T\vt x -\vt Y}^2\right] &=\EE \left[\left(\norm{\vt Y}^2-\vt Y \vt x^T \ol x^{-1} \vt x \vt Y ^T\right)\right]
        \nonumber\\
        &=\EE \left[\left(\norm{\vt Y}^2-\beta^T \ol x \beta - \vt \Noise \vt x^T \ol x^{-1} \vt x \vt \Noise ^T\right)\right]
    \end{align}
    Note that $\EE[\norm{\vt Y}^2]=\EE[\vt Y'^2]$, since $\vt Y$ and $\vt Y'$ are i.i.d.\ copies of the same random vector. Hence,
    \begin{align}
        \EE \left |    \EG(\Alg, \vt Y) \right|= \frac2n \EE \left[\vt \Noise \vt x^T \ol x^{-1} \vt x \vt \Noise ^T \right]\;,
    \end{align}
    Taking the trace of the remaining term in the expectation,
    \begin{align}
        \frac2n \EE \left[\vt \Noise \vt x^T \ol x^{-1} \vt x \vt \Noise ^T \right]  &= \frac2n \EE \left[\Tr(\vt \Noise \vt x^T \ol x^{-1} \vt x \vt \Noise ^T )\right]
        \nonumber\\
        &= \frac2n \EE \left[\Tr(\vt x^T \ol x^{-1} \vt x \vt \Noise ^T \vt \Noise )\right]
        \nonumber\\
        &=\frac2n \Tr(\vt x^T \ol x^{-1} \vt x\;\EE \left[\vt \Noise ^T \vt \Noise \right])
        \nonumber\\
        &=\frac2n\Tr(\sigma^2\vt x^T \ol x^{-1} \vt x)
        \nonumber\\
        &=\frac2n\Tr(\sigma^2 \mathbb{I}^{d\times d}) = \frac{2d\sigma^2}{n}\;.
    \end{align}
    which gives the desired result.
\end{proof}
%%% Local Variables:
%%% mode: latex
%%% TeX-master: "main"
%%% End: